\documentclass{article}

\usepackage{microtype}
\usepackage{graphicx}
\usepackage{subfigure}
\usepackage{booktabs} 
\usepackage{capt-of}  

\usepackage{hyperref}

\usepackage[accepted]{walle2}

\usepackage{changepage}
\usepackage{listings} 
\usepackage{algorithm}
\usepackage{enumitem}
\usepackage{xspace}
\usepackage{colortbl}
\definecolor{cerisepink}{rgb}{0.93, 0.23, 0.51}

\usepackage{hyperref}
\usepackage{url}
\usepackage{bm}
\usepackage{multirow}
\usepackage{wrapfig}
\usepackage{ulem}
\newcommand{\ours}{{\scshape WALL-E 2.0}\xspace}

\usepackage{amsmath}
\usepackage{amssymb}
\usepackage{mathtools}
\usepackage{amsthm}

\usepackage[capitalize,noabbrev]{cleveref}

\theoremstyle{plain}

\theoremstyle{definition}

\theoremstyle{remark}

\usepackage[textsize=tiny]{todonotes}

\icmltitlerunning{WALL-E 2.0: World Alignment by NeuroSymbolic Learning}

\begin{document}

\twocolumn[{%
\renewcommand\twocolumn[1][]{#1}%
\icmltitle{WALL-E 2.0: \underline{W}orld \underline{A}lignment by NeuroSymbo\underline{l}ic \underline{Le}arning \\improves World Model-based LLM Agents}







\begin{icmlauthorlist}
\icmlauthor{Siyu Zhou}{uts}
\icmlauthor{Tianyi Zhou}{umd}
\icmlauthor{Yijun Yang}{tencent}
\icmlauthor{Guodong Long}{uts}
\icmlauthor{Deheng Ye}{tencent}
\icmlauthor{Jing Jiang}{uts}
\icmlauthor{Chengqi Zhang}{uts}
\end{icmlauthorlist}

\icmlaffiliation{uts}{Australian AI Institute, Faculty of Engineering and IT, University of Technology Sydney, Australia}
\icmlaffiliation{umd}{Department of Computer Science, University of Maryland, College Park, USA}
\icmlaffiliation{tencent}{Tencent, China}

\icmlcorrespondingauthor{Siyu Zhou}{Siyu.Zhou-2@student.uts.edu.au}
\icmlcorrespondingauthor{Tianyi Zhou}{zhou@umiacs.umd.edu}


\icmlkeywords{LLM Agents, World Model, NeuroSymbolic}

\begin{center}
\textcolor{cerisepink}{Project: \url{https://github.com/elated-sawyer/WALL-E}}
\end{center}
\includegraphics[width=1.0 \textwidth]{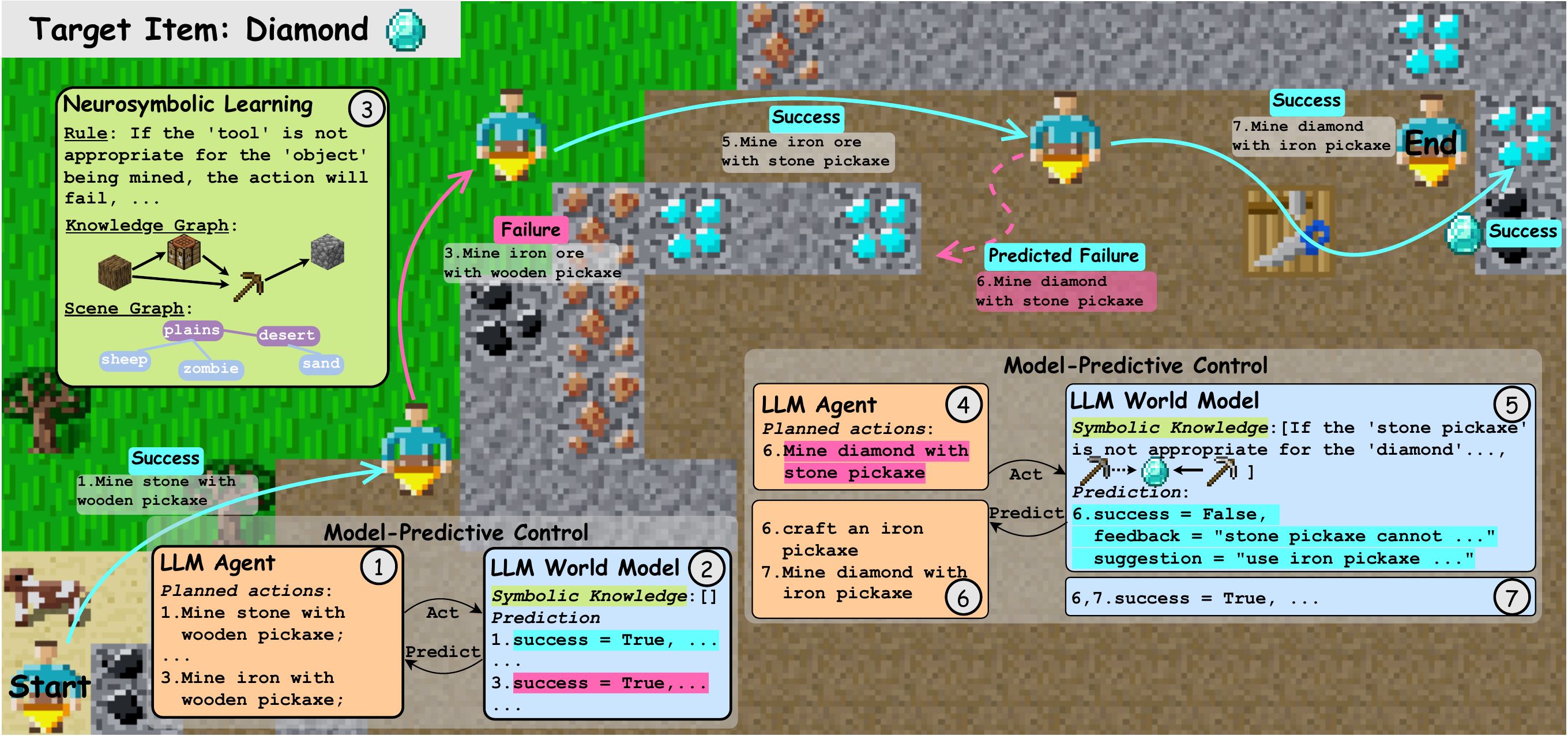} 
\vspace{-0.5em}
\captionof{figure}{
\textbf{\ours mining a diamond on Mars.}
Step 1-2: The agent makes decisions via MPC with the initial unaligned world model, resulting in a failed action for mining iron. Step 3: leveraging previous trajectories and world model predictions, \ours learns symbolic knowledge, including rules, knowledge graphs, and scene graphs.
Step 4-5: The learned symbolic knowledge helps the world model make accurate predictions and correct the previous mistake. Step 6-7: The agent adjusts its decision accordingly and replaces \textit{stone pickaxe} with \textit{iron pickaxe} toward completing the task.
\looseness-1}
\label{fig:teaser}
\vspace{1em}
}]



\printAffiliationsAndNotice{}  
\begin{abstract}
\textit{Can we build accurate world models out of large language models (LLMs)? How can world models benefit LLM agents?} The gap between the prior knowledge of LLMs and the specified environment's dynamics usually bottlenecks LLMs' performance as world models. To bridge the gap, we propose a training-free ``\textit{world alignment}'' that learns an environment's symbolic knowledge complementary to LLMs. 
The symbolic knowledge covers action rules, knowledge graphs, and scene graphs, which are extracted by LLMs from exploration trajectories and encoded into executable codes to regulate LLM agents' policies. 
We further propose an RL-free, model-based agent ``\ours'' through the model-predictive control (MPC) framework. Unlike classical MPC requiring costly optimization on the fly, we adopt an LLM agent as an efficient look-ahead optimizer of future steps' actions by interacting with the neurosymbolic world model. 
While the LLM agent's strong heuristics make it an efficient planner in MPC, the quality of its planned actions is also secured by the accurate predictions of the aligned world model. They together considerably improve learning efficiency in a new environment. 
On open-world challenges in Mars (Minecraft like) and ALFWorld (embodied indoor environments), \ours significantly outperforms existing methods, e.g., surpassing baselines in Mars by 16.1\%–51.6\% of success rate and by at least 61.7\% in score. In ALFWorld, it achieves a new record 98\% success rate after only 4 iterations.  \looseness-1
\end{abstract}

\vspace{-1.5em}
\section{Introduction}

\begin{figure}[ht]
\centering
\includegraphics[width=0.4\textwidth]{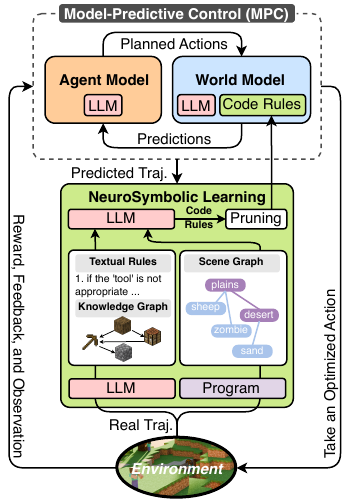}
\vspace{-0.5em}
\caption{\textbf{Overview of \ours}. The agent determines actions to take via MPC, which optimizes future steps' actions by interacting with a neurosymbolic world model. The world model adopts an LLM whose predictions are aligned with environment dynamics through code rules converted from symbolic knowledge (action rules, knowledge/scene graph) learned via inductive reasoning from real trajectories and predicted trajectories. \looseness-1
}
\label{fig:overview framework} 
\vspace{-0.8em}
\end{figure}
While large language models (LLMs) have been successfully applied to complex reasoning, generation, and planning tasks, they are not sufficiently reliable to be deployed as an agent in specific open-world environments, e.g., games, VR/AR systems, medical care, education, autonomous driving, etc~\citep{gpt-4,nips/cot, liu2024aligning}. A primary reason for the failures is the gap between the prior knowledge driven commonsense reasoning by LLMs and the specified environment's dynamics. The gap leads to incorrect predictions of future states, hallucinations, or violation of basic laws in LLM agents' decision-making process~\citep{corr/embodiedgpt,yang2024embodied,cvpr/EmbodiedQA,wu2024autohallusion}. 
Although the alignment of LLMs with human preferences has been widely studied as a major objective of LLM post-training, ``world alignment'' with an environment's dynamics has not been adequately investigated for LLM agents~\citep{hao2023reasoning,rafailov2024direct,ge2024worldgpt}. 
Moreover, many existing LLM agents are model-free: they directly generate and execute actions in real environments without being verified or optimized within a world model or simulator~\citep{corr/embodiedgpt,yao2023react,shinn2024reflexion,corr/rt-2,wu2023autogen,micheli2021language,rss/rt-1} in advance. This leads to safety risks and suboptimality of planned trajectories. \looseness-1

In this paper, we show that the ``\textbf{world alignment}'' of an LLM can significantly improve the LLM performance as a promising world model, which enables us to build more powerful embodied LLM agents in partially observable settings. 
Instead of finetuning the LLM, we introduce a training-free pipeline ``\textbf{\underline{W}orld \underline{A}lignment by NeuroSymbo\underline{l}ic \underline{Le}arning (\ours)}'' to learn symbolic knowledge that is environment-specific and complementary to the LLM's prior knowledge, by analyzing explored trajectories and predicted ones as shown in Figure~\ref{fig:overview framework}. \ours's symbolic knowledge covers action rules, knowledge graphs, and scene graphs, which can be converted into executable \textit{code rules} to turn a pretrained LLM into an accurate neurosymbolic world model (via function calling). 
It combines the strengths of both LLMs and symbolic representations in modeling environment dynamics, i.e., (1) the rich prior knowledge, probabilistic, and deductive reasoning capability of LLMs under uncertainty~\citep{hu2023language}; and (2) the formal constraints, deterministic transition rules, and environment-specific structures enforced or encoded by symbolic knowledge. In our studies, LLMs already cover the most common knowledge of dynamics but a few additional symbolic knowledge can significantly improve the world model predictions generated by LLMs. 

Different types of symbolic knowledge play important roles in building more reliable and adaptive model-based LLM agents, especially in partially observable Markov decision processes (POMDPs). 
The action rules capture and enforce deterministic constraints in the decision-making process; the knowledge graph represents feasibility constraints and action prerequisites; while the scene graph provides global information complementing the partial observations of agents. 
In \ours, we leverage LLMs' inductive reasoning capabilities to abstract concise symbolic knowledge from explored trajectories for episodes. We further develop a maximum coverage-based pruning to maintain a compact set of \textit{code rules}.
In contrast, existing LLM agents usually learn the environment dynamics through expensive finetuning of LLM policies via RL/imitation learning, or memory-heavy inference with a long input context of buffered trajectories~\citep{corr/embodiedgpt,gao2023pg-vlm,yang2024embodied,shinn2024reflexion}. 

Unlike the mainstream model-free LLM agents, \ours's precise neurosymbolic world model enables us to create more reliable and versatile model-based LLM agents for challenging open-world tasks.
Specifically, we propose a novel model-predictive control (MPC) framework for LLM agents, in which an LLM agent conducts a look-ahead optimization (planning) of future steps' actions by interacting with the world model. For example, the agent queries the world model ``What will happen if I take action $a_t$ in observation $o_t$?'', and receives the prediction with feedback/suggestions according to the \textit{code rules}, based on which the agent chooses to execute the plan or keep refining it until it passes the world model's examination, i.e., not violating any constraints of the environments and leading the agent to preferred (predicted) states. 
Our \textbf{LLM-based MPC} framework overcomes the inefficiency of classical MPC that requires online $k$-step optimization, by exploiting the strong reasoning and instruction following capability of LLMs. 

We evaluate \ours in challenging open-worlds such as Mars (Minecraft-like) and ALFWorld (embodied indoor environment) where the agents can explore freely and target complicated tasks. Our main contributions can be summarized as:\looseness-1
\begin{itemize}[leftmargin=1em, itemsep=0.1em]
\vspace{-0.5em}
    \item We address the ``world alignment'' challenge for LLM agents and propose a novel training-free approach ``\ours'' to align LLMs with environment dynamics, resulting in neurosymbolic world models. 
    \item We develop a novel LLM-based MPC framework for world model-based LLM agents. 
    \item \ours achieves state-of-the-art performance in Mars and ALFWorld.  
\end{itemize}

\paragraph{\textcolor{cerisepink}{New Update to WALL-E 1.0}~\citep{zhou2024wall}}  
\ours introduces several key enhancements and new features that extend the original WALL-E framework with improved planning capability and adaptability:
\begin{itemize}[leftmargin=1em, itemsep=0.1em]
    \item \textbf{Inductive Learning of Knowledge Graph}: \ours constructs a knowledge graph by performing inductive reasoning of LLM to infer symbolic relations (e.g., require, consume) from past experience, enriching the agent’s understanding of action preconditions and effects.
    \item \textbf{Dynamic Scene Graph Extraction}: \ours dynamically build a scene graph from real-time environment feedback, providing a structured and up-to-date representation of objects and their spatial relationships in the environment.
    \item \textbf{NeuroSymbolic World Model Integration}: \ours incorporates executable action rules, knowledge graph, and scene graph with an LLM, resulting in a unified neurosymbolic world model. This allows the LLM agent to perform scene-aware, structured, and interpretable planning, which significantly improves the agent's adaptation to complex, dynamic environments.
\end{itemize}

\section{Related Work}

Due to limited space, we provide a more comprehensive related work in Appendix \ref{sec:Detailed Related Work}. Although the alignment of LLMs has been widely studied as a major objective of LLM post-training~\citep{yang2023enabling, Zhu2023Large, mu2023can, yang2023failures, luo2023chatrule}, ``world alignment'' with an environment's dynamics has not been adequately investigated for LLM agents. To bridge this gap, we propose a training-free method that learns an environment's symbolic knowledge complementary to LLMs and uses it to steer a pretrained LLM toward an accurate neurosymbolic world model. 
Unlike previous methods that often require expensive RL/imitation learning for LLM policy finetuning, or rely on memory-intensive inference over large buffers of past trajectories, or build explicit world models entirely from scratch \citep{corr/embodiedgpt, gao2023pg-vlm, yang2024embodied, shinn2024reflexion, tang2024worldcoder}, our approach focuses on selectively aligning the pretrained LLM to new environment dynamics without heavy overhead. By synthesizing symbolic insights and integrating them into the LLM’s capabilities, \ours creates a more efficient and reliable model-based agent for open-world tasks.

\section{World Alignment by NeuroSymbolic Learning
(\ours)}

We consider a Partially Observable Markov Decision Process (POMDP) and denote it by the tuple 
$\mathcal{M} = \bigl(\mathcal{S}, \mathcal{A}, \mathcal{O}, r, \mathcal{T}, \gamma\bigr)$, in which $\mathcal{S}$ represents the set of hidden (unobservable) states of the environment, $\mathcal{A}$ is the set of actions, 
$\mathcal{O}$ denotes to the space of textual observations available to the agent, including possibly partial or noisy descriptions of the environment,
$r: \mathcal{S} \times \mathcal{A} \to \mathbb{R}$ defines a reward of taking an action $a$ in a given state $s$, $\mathcal{T}(s_{t+1} \mid s_t, a_t)$ is the transition probability conditioned on state $s_t$ and action $a_t$ that determines the transition probability of the process to state $s_{t+1}$, and $\gamma \in [0,1]$ is the discount factor. Due to the partially observable setting of the process, the policy $\pi$ does not have access to the hidden state $s_t$ but instead receives an observation $o_t \in \mathcal{O}$ at the time step $t$. \looseness-1

\subsection{NeuroSymbolic Learning of Code Rules}
\label{sec:NeuroSymbolic Learning}

\begin{figure*}[t!]
\begin{center}
\includegraphics[width=0.95 \linewidth]{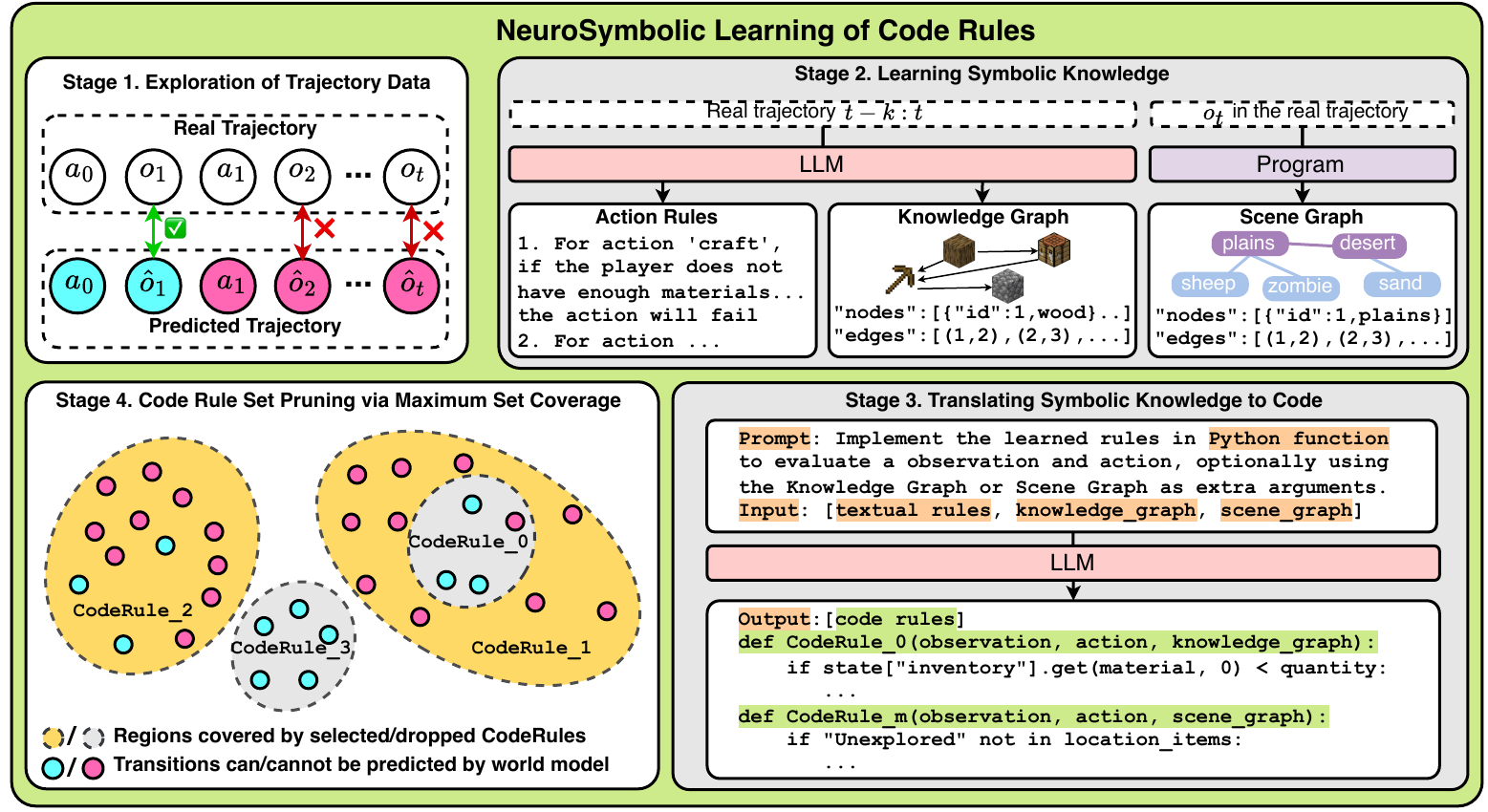} 
\end{center}
\vspace{-1em}
\caption{\textbf{NeuroSymbolic Learning of Code Rules}.
\ours iteratively refines the symbolic knowledge with the agent's actual trajectories in the environment and the world model's predicted trajectories. 
The NeuroSymbolic learning takes 4 stages: (1) comparing predicted and actual trajectories; (2) learning new symbolic knowledge from real trajectories; (4) translating symbolic knowledge to code; and (4) Code rule set pruning via solving a maximum coverage problem. }
\label{fig:Rule Learning Framework} 
\end{figure*}

\subsubsection{Stage 1: Exploration of Trajectory Data}
\label{sec:Trajectories Data Processing}

To evaluate inconsistencies between the LLM world model and the real environment, we compare the environment-generated (real) trajectory $\tau^{\text{real}} = \{ \delta^{\text{real}} = (o_t, a_t, o_{t+1}) \}_{t=0}^{T}$ and the model-predicted trajectory $\tau^{\text{predicted}} = \{ \delta^{\text{predicted}} = (o_t, a_t, \hat{o}_{t+1}) \}_{t=0}^{T}$. 

Specifically, we simplify the world model’s prediction task to a binary classification of whether a transition will succeed or fail, instead of directly predicting the next observation $o_{t+1}$. This design choice is motivated by the fact that $o_{t+1}$ can be derived from the success or failure of an action. 
For example, if the action “craft a table” succeeds, the number of tables increases and the corresponding materials decrease in $o_{t+1}$. In contrast, directly predicting the full observation $o_{t+1}$ is more complex and susceptible to noise from stochastic environmental factors (e.g., weather or creature presence). Thus, we perform consistency checks based on transition success/failure only.

We then classify all transitions in $\tau^{\text{predicted}}$ into two sets: correctly predicted transitions $\delta^{\text{cor}} \in \mathcal{D}^{\text{cor}}$ and incorrectly predicted transitions $\delta^{\text{inc}} \in \mathcal{D}^{\text{inc}}$, based on whether the predicted success/failure matches the actual outcome.
As illustrated in Step 1 of Figure~\ref{fig:Rule Learning Framework}, the real trajectories are used to induce new code rules that align the LLM world model more closely with the actual environment dynamics. Meanwhile, $\mathcal{D}^{\text{incorrect}}$ serves as a diagnostic set to evaluate and prune redundant or conflicting code rules via a maximum coverage formulation (see Sec.~\ref{sec:rule pruning} for details).

\subsubsection{Stage 2: Learning Symbolic Knowledge}
\label{sec:Learning Symbolic knowledge from Real Trajectories.}

\paragraph{Extraction of Action Rules: Deterministic Transitions}
We first leverage the inductive reasoning capability of LLMs to extract action rules from real historical trajectories $\tau^\text{real}_{t-k}$, where $\tau^\text{real}_{t-k}$ denotes the past $k$ real transitions before timestep $t$.
The parameter $k$ defines a finite context window for rule extraction.
The extracted action rules capture deterministic constraints in the environment and are used to guide and enforce correct decision-making. For example, an action rule might specify: ``For action \textit{make}, if \textit{table} is not in \textit{near\_objects}, the action will fail.''
Formally,

\begin{equation}
\mathcal{R}_{t} = f_\text{InductiveReasoning}\Bigl(\tau^\text{real}_{t-k}\Bigr)\cup \mathcal{R}_{t-1}
\end{equation}
where $f_\text{InductiveReasoning}$ is implemented by prompting an LLM to generate textual rules (see Appendix \ref{sec:Prompt for Learn New Rules from Real Trajectories} for detailed prompts and stage 2 in Fig.~\ref{fig:Rule Learning Framework} for visualized examples),
and $\mathcal{R} = \{r_1, r_2, \ldots, r_{|\mathcal{R}|}\}$ denotes the extracted rule set.

\paragraph{Knowledge Graph: Constraints in the POMDP}

Next, we construct a Knowledge Graph (KG) based on exploratory trajectories to represent environments' feasibility constraints and actions' prerequisites. 
Specifically, let
\begin{equation}
\mathcal{G}^{\text{knowledge}} = (\mathcal{V}, \mathcal{E})
\end{equation}
be a directed, labeled graph where $\mathcal{V}$ is the set of entities (e.g., items, materials, or location-specific objects) and $\mathcal{E}$ contains edges that encode constraints. An edge $(u, v, \ell) \in \mathcal{E}$ with label $\ell$ captures a constraint on how an entity $u$ (e.g., a material in Mars) relates to another entity $v$ (e.g., a product built upon the material) under the label $\ell$, e.g., ``require'', ``consume'', and ``enable''. 
For instance, if we observe that item $A$ always fails to be created unless $x$ units of $B$ and $y$ units of $C$ are available, we add the following edges $(B, A, \text{require } x)$ and $(C, A, \text{require } y)$ to the KG.

\paragraph{Scene Graph: Complementing Partial Observations}
To address the challenges posed by the partial observability of POMDP, we build a Scene Graph (SG) to record global information and complement the agent’s immediate observation $o_t$ at time step $t$. Formally,
\begin{equation}
\mathcal{G}_{t}^{\text{scene}} = (\mathcal{V}_{t}, \mathcal{E}_{t})
\end{equation}
where $\mathcal{V}_{t}$ denotes entities, objects, or locations observed by the agent interacting with the environment, while $\mathcal{E}_{t}$ indicates spatial relationships such as ``ownership'', ``located in'' or ``adjacency''. For example, if the environment contains multiple rooms, and each room includes certain items, SG can record the specific items located in every room, complementing the agent’s local observation with global information.
At each time step $t$, we update the SG via:
\begin{equation}
\mathcal{G}_{t}^{\text{scene}} = f_{\mathrm{scene}}(o_t)\cup\mathcal{G}_{t-1}^{\text{scene}},
\end{equation}
where $f_{\mathrm{scene}}$ is a function that analyzes the agent's observation $o_{t}$ and generates a sub-graph recording the spatial relationships among entities present in $o_{t}$.

\begin{algorithm}[t]
\caption{Greedy Algorithm for the Maximum Coverage Problem in Eq.~(\ref{eq:max_coverage})}
\label{alg:Greedy Algorithm}
\begin{algorithmic}[1]
\STATE \textbf{Input:} $\mathcal{D}^{\text{inc}} = \{ \delta^{\text{inc}}_{1}, \delta^{\text{inc}}_{2}, .., \delta^{\text{inc}}_{N} \}$, 
$\mathcal R^{\text{Code}} = \{ r^{\text{Code}}_1, r^{\text{Code}}_2, .., r^{\text{Code}}_M \}$, 
$a_{ij}$: Indicator matrix where $a_{ij} = 1$ if $\delta^{\text{inc}}_{j} \in r^{\text{Code}}_{i}$, otherwise $a_{ij} = 0$

\STATE \textbf{Initialize} $\mathcal R^{*}\gets \emptyset$, $\mathcal{D}^{cov}\gets \emptyset$

\WHILE{$\mathcal{D}^{cov} \neq \mathcal{D}^{\text{inc}}$}
    \STATE For each rule $r^{\text{Code}}_{i} \in \mathcal R^{\text{Code}}$, compute:
    \begin{align*}
        &\text{gain}(r^{\text{Code}}_{i}) = \\
        &\left|\mathcal{D}^{cov} \cup \{ \delta^{\text{inc}}_{j}\in\mathcal{D}^{\text{inc}} : a_{ij} = 1 \} \right|
        - |\mathcal{D}^{cov}| 
    \end{align*}
    
    \STATE Get the index of $r^{\text{Code}}_{i}$ with the largest gain, i.e.,
    \begin{equation*}
        i^* \gets \arg\max \text{gain}(r^{\text{Code}}_{i})
    \end{equation*}
    
    \IF{$\text{gain}(r^{\text{Code}}_{i^*}) = 0$}
        \STATE \textbf{Break} \COMMENT{Terminate if no $r^{\text{Code}}_{i}$ can cover any additional $\delta^{\text{inc}}$ }
    \ENDIF

    \STATE Add $r^{\text{Code}}_{i^*}$ to the selected rules set:
    \begin{equation*}
        \mathcal R^{*} \gets \mathcal R^{*} \cup \{ r^{\text{Code}}_{i^*} \}
    \end{equation*}
    
    \STATE Update the covered set:
    \begin{equation*}
        \mathcal{D}^{cov} \gets \mathcal{D}^{cov} \cup \{ \delta^{\text{inc}}_{j}\in\mathcal{D}^{\text{inc}} : a_{i^*j} = 1 \}
    \end{equation*}
    
    \IF{$|\mathcal R^{*}| = l$}
        \STATE \textbf{Break} \COMMENT{Terminate if hit the limit $l$}
    \ENDIF

\ENDWHILE
\STATE \textbf{Output:} Set of selected rules $\mathcal R^{*}$
\end{algorithmic}
\end{algorithm}

\subsubsection{Stage 3: Translating Symbolic Knowledge to Compact Code Rules}
\label{sec:Translating Symbolic Knowledge to Code}

Once we obtain symbolic knowledge—including action rules $\mathcal{R}$, knowledge graphs (KG) $\mathcal{G}^{\text{knowledge}}$, and scene graphs (SG) $\mathcal{G}^{\text{scene}}$—in stage 2, we operationalize this knowledge as executable code rules,
enabling symbolic reasoning over observation-action pairs. Specifically, we leverage the coding capabilities of LLMs to translate the symbolic structures into Python functions that determine whether a given observation-action pair conforms to previously acquired symbolic constraints. These functions return a boolean outcome along with structured feedback messages (see Appendix~\ref{sec:Translate Natural Language Rules to Code} for more details). $\mathcal{R}^{\text{Code}}$ denotes the set of executable code rules. Each individual rule $r^{\text{code}}_{m} \in \mathcal{R}^{\text{Code}}$ follows the format defined below:\looseness-1
\begin{lstlisting}[language=Python, frame=none, aboveskip=0pt, belowskip=0pt, basicstyle=\small\ttfamily]
def CodeRule_m(obs, action, KG/SG):
    ...
    return feedback, suggestion, flag
\end{lstlisting}
in which \texttt{flag} is a boolean value equal to \texttt{True} if the action succeeds under the observation, \texttt{False} otherwise; \texttt{feedback} and \texttt{suggestion} are procedurally generated string messages regarding the action's outcome and providing guidance if \texttt{flag} is \texttt{False}.
In theory, these code rules can embed a variety of symbolic knowledge into compact, executable functions, reducing complexity and enabling efficient decision-making processes.

\subsubsection{Stage 4: Code Rules Pruning}
\label{sec:rule pruning}

As the agent accumulates more interaction data over time, the number of learned code rules in Stage 3 naturally grows. To prevent the rule set \(\mathcal{R}^{\text{code}} = \{r^{\text{code}}_m\}_{m=0}^{M}\) from becoming unnecessarily large or redundant, we introduce an effective pruning strategy based on solving a maximum set coverage problem, ensuring that the retained rules are both minimal and impactful.

As discussed earlier, we simplify the LLM world model to predict only whether a given transition will succeed or fail, which is sufficient for accurately determining the future state \(o_{t+1}\). Accordingly, the learned rules are designed to evaluate transition feasibility. Each rule follows a unified structure: “if [state conditions] then [transition success/failure], otherwise [null].”
A rule is activated when the current state satisfies its conditions, producing a non-null output. If the output of an activated rule \( r^{\text{code}}_i \) corrects an initially incorrect prediction for a transition \( \delta^{\text{inc}}_j \), we say that \( \delta^{\text{inc}}_j \) is covered by \( r^{\text{code}}_i \). Based on this notion of coverage, we formulate the rule selection process as a maximum coverage optimization problem, aiming to retain the smallest set of rules that collectively correct the largest number of incorrect predictions. (See Appendix~\ref{sec:Learned Rules} for examples of learned rules.)
\looseness-1

\begin{align}\label{eq:max_coverage}
    \mathcal{R}^{*} = \arg\max_{\mathcal{R}\subseteq \mathcal{R}^{\text{Code}}} |\bigcup_{r^{\text{Code}}\in \mathcal{R}} \mathcal{D}^{cov}|, \text{ s.t., } |\mathcal{R}|\leq l.
\end{align}
where $\mathcal D^{cov}$ is the subset of transitions covered by rules, i.e., $\mathcal D^{cov}\triangleq\{\delta^{\text{inc}}\in \mathcal D^{\text{inc}}: \delta^{\text{inc}}~~\text{covered by}~~r^{\text{Code}}\in \mathcal{R}^{\text{Code}}\}$.
The parameter $l > 0$ is the limit of selected rules, and we find that a large $l$ leads to better performance.
Our goal is to select a subset of rules $\mathcal{R}^{*} \subseteq \mathcal{R}^{\text{Code}}$ that maximizes coverage of $\mathcal{D}^{\text{inc}}$.
We solve this problem using a greedy method in Algorithm~\ref{alg:Greedy Algorithm}.
Through this process, we eliminate \textbf{rules covering only correct transitions}, as they do not address misalignments, and \textbf{redundant rules} fully covered by more comprehensive ones (see Step 5 of rule learning in Figure \ref{fig:Rule Learning Framework}).

The NeuroSymbolic learning process consists of four stages: (1) comparing predicted and real trajectories, (2) extracting symbolic knowledge, (3) translating it into executable code rules, and (4) pruning the rule set via a maximum coverage objective. This procedure, defined as \(\textsc{NSLearning}()\), yields a compact and effective rule set:
\begin{equation}
    \mathcal{R}_{t}^{\text{code}} \gets \textsc{NSLearning}(\tau^{\text{pred}}, \tau^{\text{real}}).
\end{equation}
The resulting rules enhance the LLM world model’s alignment and improve decision-making quality.

\begin{algorithm}[t]
\caption{Model-Predictive Control (MPC)}
\label{alg:MPCPlanner}
\begin{algorithmic}[1]
\STATE \textbf{Input:} $o_t$, $\mathcal{R}_t^{\text{code}}$
\STATE \textbf{Initialize:} $feedback \gets []$, $sugg \gets []$, $replan\_count \gets 0$

\REPEAT
    \STATE $a_t \gets \textsc{LLMAgent}(o_t, feedback, sugg)$
    \STATE $\hat{o}_{t+1}, feedback, sugg, flag \gets \textsc{MapExecute}(\mathcal{R}_t^{\text{code}}, \textsc{WM}(o_t, a_t))$
    \STATE $replan\_count \gets replan\_count + 1$
    \IF{flag}
        \STATE \textbf{break} /*Action accepted*/
    \ENDIF
\UNTIL{$replan\_count \geq \textsc{ReplanLimit}$}
\STATE \textbf{Output:} Planned action $a_t$, predicted outcome $\hat{o}_{t+1}$
\end{algorithmic}
\end{algorithm}

\begin{algorithm}[t]
\caption{\textsc{WALL-E 2.0}}
\label{alg:overallpipeline}
\begin{algorithmic}[1]
\STATE \textbf{/*} \textsc{NSLearning()} is detailed in Section~\ref{sec:NeuroSymbolic Learning}; \textsc{MPC()} is described in Section~\ref{sec:MPCinWALLE} \textbf{*/}
\STATE \textbf{Initialize} $\mathcal{R}^{\text{code}} \gets \emptyset$, $\tau^{\text{real}} \gets []$, $\tau^{\text{pred}} \gets []$, $t \gets 0$, $o_0 \gets \textsc{env}()$
\WHILE{$\neg \textsc{taskcomplete}(o_t)$ \textbf{and} $\neg \textsc{agentdied}(o_t)$}
    \STATE $a_t, \hat{o}_{t+1} \gets \textsc{MPC}(o_t, \mathcal{R}_t^{\text{code}})$
    \STATE $o_{t+1} \gets \textsc{env}(a_t)$
    \STATE $\tau^{\text{real}}.\textsc{append}((o_t, a_t, o_{t+1}))$
    \STATE $\tau^{\text{pred}}.\textsc{append}((o_t, a_t, \hat{o}_{t+1}))$
    \STATE $\mathcal{R}_{t}^{\text{code}} \gets \textsc{NSLearning}(\tau^{\text{pred}}, \tau^{\text{real}})$ 
    \STATE $t \gets t + 1$
\ENDWHILE
\end{algorithmic}
\end{algorithm}

\subsection{Model-Predictive Control for World Model-based LLM Agents}
\label{sec:MPCinWALLE}

As illustrated in Figure~\ref{fig:overview framework}, the LLM-based agent operates within a POMDP, using an LLM-augmented world model combined with executable code rules for planning and action selection. Specifically, at each decision step, the agent queries the world model, asking, ``What will happen if action \(a_t\) is taken given the current observation \(o_t\)?''. The world model predicts the subsequent observation \(\hat{o}_{t+1}\) and uses code rules to provide structured feedback and suggestions. This information guides the agent's decision-making, allowing iterative refinement of its action plan until it aligns fully with the environment's constraints and moves toward desirable states. This process is detailed in Algorithm~\ref{alg:MPCPlanner}.

Crucially, unlike conventional MPC methods that rely heavily on random action sampling, our approach leverages the pretrained LLM’s rich priors and reasoning capabilities, making it a highly efficient look-ahead optimizer. This substantially reduces the search space and computational complexity, especially beneficial in complex or high-dimensional environments.

Formally, after the LLM world model initially predicts \(\hat{o}_{t+1} = \mathrm{WM}(o_{t}, a_{t})\), the code rules \(\mathcal{R}^{\text{code}}\) verify the prediction's consistency: 
\begin{align}
    &\hat{o}_{t+1},\; \text{feedback},\; \text{sugg},\; \text{flag} \\
    &= \textsc{MapExecute}\bigl(\mathcal{R}^{\text{code}},\; \mathrm{WM}(o_{t}, a_{t})\bigr), \nonumber
\end{align}
where \(\textsc{MapExecute}(\cdot)\) checks whether the activated code rules agree with the LLM prediction. If discrepancies arise, the outputs of the code rules override the predictions of the LLM world model, ensuring consistency with the environment's dynamics. In addition, as described in Section~\ref{sec:Translating Symbolic Knowledge to Code}, activated rules provide structured feedback, explicit suggestions and flag signal to guide action adjustment in accordance with known constraints. This feedback is integrated into the agent’s planning process, allowing it to revise its decision as needed. Specifically, the next action is generated as \looseness-1 
\begin{equation}
    a_t \gets \textsc{LLMAgent}(o_t, feedback, sugg).
\end{equation}
\ours operates in an iterative loop that alternates between \textbf{MPC} and \textbf{NeuroSymbolic learning}. During each MPC phase, actions proposed by the LLM agent are validated by the world model using the current symbolic knowledge. In parallel, new symbolic knowledge is continuously extracted from recent interactions, converted into executable rules, and pruned before being appended to the world model. This loop continues iteratively, enabling efficient adaptation without training.  The overall process is detailed in Algorithm~\ref{alg:overallpipeline}.

\section{Experiments}
\label{Experiments}

\begin{table*}[t]
\vspace{-.9em}
\caption{Comparison of \ours with RL-based (PPO~\citep{schulman2017proximal} \& DreamerV3~\citep{hafner2023mastering}) and LLM-based methods (ReAct~\citep{yao2022react}, Reflexion~\citep{shinn2023reflexion}, Skill Library~\citep{wang2024jarvis}, IfR~\citep{tang2024mars}, WALL-E 1.0~\citep{zhou2024wall}) in Mars~\cite{tang2024mars}. LLM-based methods and RL-based methods' results are averaged over 9 trials and 20 trials, respectively ($*$-reported in previous work). 
RL trains a policy separately for each world and is supposed to be better than LLM-based. 
Reward are accumulated and reflects unlocked achievements. 
Score (\%) is the weighted geometric mean of success rates, emphasizing rare and challenging achievements. 
The best results are in \textbf{bold}.
\textbf{\ours outperforms other baselines even in counter-commonsense scenarios} that conflict with the LLM’s prior knowledge, exhibiting superior planning ability and robust adaptability.\looseness-1}
\vspace{-1.5em}
\label{tab:mars-main-result}
\vskip 0.15in
\begin{center}
\begin{small}
\begin{sc}
\renewcommand\arraystretch{0.65}
\resizebox{1\linewidth}{!}{
    \begin{tabular}{llcccccccc}
    \toprule
    \multirow{2}{*}{\textbf{Metrics}}  & \multirow{2}{*}{\textbf{Mod. Type}\footnotemark}  & \multicolumn{2}{c}{\textbf{RL-based methods}} & \multicolumn{6}{c}{\textbf{LLM-based methods}} \\ \cmidrule(lr){3-4}  \cmidrule(lr){5-10} 
      &   & \textbf{PPO*} & \textbf{DreamerV3*} & \textbf{ReAct*} & \textbf{Reflexion*} & \textbf{Skill Library*} & \textbf{IfR*} & \textbf{WALL-E 1.0} & \textbf{\ours} \\
    \midrule
    \multirow{9}{*}{\textbf{Reward} $\uparrow$ } 
    & Default      & $1.9 \pm 1.4$ & $11.5 \pm 1.6$ & $7.7 \pm 1.6$ & $6.0 \pm 1.7$ & $8.0 \pm 2.1$ & $9.0 \pm 2.3$ & $9.4 \pm 2.4$ & $\bm{9.5 \pm 2.1}$ \\
    & Terrain      & $-0.1 \pm 0.6$ & $9.3 \pm 2.2$ & $7.4 \pm 2.7$ & $6.4 \pm 3.0$ & $9.5 \pm 2.9$ & $8.0 \pm 3.7$ & $9.8 \pm 2.9$ & $\bm{10.7 \pm 2.6}$ \\
    & Survival     & $-0.6 \pm 0.5$ & $8.6 \pm 4.1$ & $6.4 \pm 3.7$ & $4.6 \pm 3.9$ & $7.9 \pm 2.9$ & $7.7 \pm 3.7$ & $10.5 \pm 5.0$ & $\bm{13.8 \pm 4.4}$ \\
    & Task. Dep.   & $2.1 \pm 1.2$ & $8.8 \pm 2.8$ & $5.0 \pm 2.1$ & $3.2 \pm 1.6$ & $1.5 \pm 1.9$ & $5.6 \pm 2.9$ & $3.9 \pm 2.0$ & $\bm{6.4 \pm 2.9}$ \\
    & Terr. Surv.  & $0.0 \pm 0.7$ & $7.1 \pm 2.1$ & $6.7 \pm 2.5$ & $4.9 \pm 2.5$ & $3.0 \pm 2.5$ & $\bm{6.8 \pm 1.9}$ &  $4.1 \pm 1.9$ & $5.5 \pm 2.7$ \\
    & Terr. Task.  & $-0.7 \pm 0.3$ & $6.6 \pm 0.7$& $4.8 \pm 2.0$ & $5.3 \pm 2.5$ & $5.5 \pm 1.5$ & $\bm{6.9 \pm 1.8}$ & $3.1 \pm 2.1$ & $5.8 \pm 2.2$ \\
    & Surv. Task.  & $-0.6 \pm 0.4$ & $9.6 \pm 3.4$ & $1.5 \pm 1.3$ & $1.0 \pm 1.6$ & $2.3 \pm 1.5$ & $3.3 \pm 1.4$ & $\bm{3.6 \pm 1.3}$ & $3.2 \pm 1.4$ \\
    & All three    & $0.1 \pm 0.8$ & $5.1 \pm 1.8$ & $0.7 \pm 1.6$ & $-0.4 \pm 0.7$ & $-0.5 \pm 0.5$ & $0.1 \pm 0.5$ & $0.8 \pm 0.4$ & $\bm{1.3 \pm 1.6}$ \\ \cmidrule(lr){2-10} 
    & Average & $0.0$ & $7.9$ & $4.6$ & $3.6$ & $4.2$ & $5.5$ & $5.1$ & $\bm{6.7}$ \\
    \midrule
    \multirow{9}{*}{\textbf{Score (\%)} $\uparrow$ } 
    & Default      & $1.3 \pm 1.7$ & $14.2 \pm 1.3$ & $8.0 \pm 1.5$ & $5.3 \pm 0.9$ & $8.3 \pm 1.3$ & $13.0 \pm 2.1$ & $17.6 \pm 1.5$ & $\bm{20.3 \pm 1.8}$ \\
    & Terrain      & $0.3 \pm 0.1$ & $13.0 \pm 1.6$ & $7.6 \pm 2.6$ & $7.4 \pm 1.6$ & $11.9 \pm 3.4$ & $11.8 \pm 2.9$ & $18.0 \pm 1.7$ & $\bm{27.8 \pm 1.7}$ \\
    & Survival     & $0.2 \pm 0.0$ & $10.8 \pm 2.8$ & $8.0 \pm 0.6$ & $5.5 \pm 1.7$ & $9.7 \pm 2.0$ & $11.0 \pm 3.7$ & $21.8 \pm 1.8$ & $\bm{50.8 \pm 1.1}$ \\
    & Task. Dep.   & $1.7 \pm 0.6$ & $12.1 \pm 1.9$ & $4.6 \pm 1.6$ & $2.2 \pm 0.8$ & $1.5 \pm 0.6$ & $6.9 \pm 2.5$ & $4.8 \pm 2.2$ & $\bm{9.3 \pm 2.0}$ \\
    & Terr. Surv.  & $0.4 \pm 0.1$ & $7.9 \pm 1.3$ & $7.1 \pm 3.0$ & $4.7 \pm 1.6$ & $2.8 \pm 0.6$ & $6.7 \pm 0.8$ & $6.9 \pm 1.5$ & $\bm{8.6\pm 1.9}$ \\
    & Terr. Task.  & $0.1 \pm 0.1$ & $4.2 \pm 0.1$ & $3.8 \pm 0.3$ & $5.5 \pm 1.7$ & $4.1 \pm 0.7$ & $\bm{7.1 \pm 2.5}$ & $2.7 \pm 1.1$ & $4.7 \pm 2.0$ \\
    & Surv. Task.  & $0.1 \pm 0.1$ & $15.9 \pm 2.6$ & $1.3 \pm 0.2$ & $1.1 \pm 0.1$ & $1.9 \pm 0.1$ & $2.1 \pm 0.4$ & $\bm{3.5 \pm 1.5}$ & $3.3 \pm 1.9$ \\
    & All three    & $0.6 \pm 0.2$ & $4.0 \pm 0.3$ & $1.0 \pm 0.3$ & $0.2 \pm 0.1$ & $0.2 \pm 0.0$ & $0.6 \pm 0.0$ & $0.5 \pm 0.3$ & $\bm{2.2 \pm 1.6}$ \\ \cmidrule(lr){2-10} 
    & Average & $0.6$ & $10.3$ & $5.2$ & $4.0$ & $5.0$ & $7.4$ & $8.3$ & $\bm{15.3}$ \\
    \bottomrule
    \end{tabular}
    }
\end{sc}
\end{small}
\end{center}
\vskip -0.2in
\end{table*}

\subsection{Benchmarks}
\begin{itemize}[leftmargin=1em, itemsep=0.0em]
\item \textbf{Mars}~\citep{tang2024mars} is an interactive environment designed to benchmark models’ ability to perform situated inductive reasoning—deriving and applying new rules in unfamiliar contexts. Built on Crafter~\citep{hafner2021benchmarking}, Mars modifies terrain, survival settings, and task dependencies to generate diverse worlds. Agents must interact with these worlds, adapting to novel rules rather than relying on pre-existing knowledge like “consuming cows increases health.” This makes Mars a unique testbed for evaluating agent adaptability and reasoning in dynamic environments.
\item \textbf{ALFWorld}~\citep{shridhar2020alfworld} is a virtual environment designed as a visual/text-based simulation where agents perform tasks by interacting with a simulated household environment. 
Text-based refers to scenarios without visual modality, where the agent receives ground-truth information directly as part of the input. In contrast, Visual-based scenarios involve visual perception, where ground-truth information is not provided explicitly. Instead, agents must interpret raw visual inputs using perception tools to understand the environment.
This benchmark includes six distinct task types, each requiring the agent to accomplish a high-level objective, such as placing a cooled lettuce on a countertop. See Appendix~\ref{sec:Experiment Details} for details. 
\looseness-1
\end{itemize}

\subsection{Evaluation Metrics}
\begin{itemize}[leftmargin=1em, itemsep=0.1em]
\item \textbf{Mars:}
(1) \textbf{Reward} (higher is better): the sum of sparse rewards given during an episode, including $+1$ for unlocking achievements, $-0.1$ for health points lost, and $+0.1$ for health points regained. The reward primarily reflects the number of achievements unlocked. 
(2) \textbf{Score} (higher is better): a single aggregated value derived from the geometric mean of success rates across achievements, weighing rare and difficult achievements more heavily to better reflect overall capability. 
\item \textbf{ALFWorld:}
(1) \textbf{Success rate} (higher is better): the percentage of tasks the agent completes successfully. 
\end{itemize}

\subsection{Main Results}
\vspace{-.5em}

\footnotetext[1]{World types include Default (original Crafter setting with no modifications), individual modifications (Terrain, Survival, Task Dependency), and combinations of two or all three modifications (Terr. Surv., Terr. Task., Surv. Task., All Three).}
\begin{table*}[t]
\caption{Comparison of \ours and baselines on 134 testing tasks from the ALFWorld~\cite{shridhar2020alfworld}. $*$-reported in previous work. 
The highest success rate (\%) for each task is highlighted in \textbf{bold}. 
\textbf{\ours outperforms all other baselines.}
\looseness-1}
\vspace{-0.3em}
\label{tab:alfworld-main-results}
\begin{center}
\begin{small}
\begin{sc}
\renewcommand\arraystretch{0.65}
\resizebox{0.9\linewidth}{!}{

    \begin{tabular}{llccccccc}
    \toprule
    \multicolumn{2}{l}{\multirow{2}{*}{\textbf{Method}}}     & \multicolumn{7}{c}{\textbf{Success Rate (\%) $\uparrow$}} \\ \cmidrule(lr){3-9} 
    \multicolumn{2}{l}{}                                                                        & Avg.      & Pick      & Clean     & Heat      & Cool      & Examine   & Picktwo \\ \midrule
    \multirow{5}{*}{\rotatebox{90}{\textbf{VLMs}}} & MiniGPT-4*~\citep{zhu2023minigpt}          &16           &4           &0           &19           &17           &67           &6           \\
                                                   & BLIP-2*~\citep{li2023blip}                 &4           &0           &  6         &  4         &  11         &  6         &  0         \\
                                                   & LLaMA-Adapter*~\citep{gao2023llama}        & 13          & 17          & 10          & 27          & 22          & 0          &  0         \\
                                                   & InstructBLIP*~\citep{dai2023instructblip}  & 22          & 50          & 26          & 23          & 6          &  17         &  0         \\
                                                   & EMMA*~\citep{yang2024embodied}             & \textbf{82}  & \textbf{71}  & \textbf{94}  & \textbf{85} & \textbf{83}  & \textbf{88} &  \textbf{67}         \\ \midrule
    \multirow{9}{*}{\rotatebox{90}{\textbf{LLMs}}} & BUTLER*~\citep{micheli2021language}        & 26          & 31          & 41          & 60          & 27          & 12          &  29         \\
                                                   & GPT-BUTLER*~\citep{micheli2021language}    & 69          & 62          & 81          & 85          & 78          & 50          &  47         \\
                                                   & DEPS~\citep{wang2023describe}              & 76          & 93          & 50          &  80         & \textbf{100}          & \textbf{100}          & 0          \\
                                                   & AutoGen*~\citep{wu2023autogen}             & 77          & 92          & 74          &  78         & 86          &  83         &  41         \\
                                                   & ReAct~\citep{yao2023react}                 & 74          & 79           & 54           & 96          & 85        & 83        & 51        \\
                                                   & AdaPlanner~\citep{sun2024adaplanner}       & 91          & \textbf{100} & \textbf{100} & 89           & \textbf{100} & 97           & 47           \\
                                                   & Reflexion~\citep{shinn2024reflexion}       & 86          & 92           & 94           & 70           & 81           & 90           & 88           \\
                                                   & RAFA~\citep{liu2023reason}                 & 95 & \textbf{100} & 97           & 91           & 95           & \textbf{100} & 82           \\
                                                   & WALL-E 1.0~\citep{zhou2024wall}        &  95 & \textbf{100} & 97 & \textbf{100}         & 86    & 85   & \textbf{100} \\ 

                                                   & \textbf{\ours (ours)}                     &  \textbf{98} & \textbf{100} & \textbf{100} & 96         & \textbf{100}    & \textbf{100}   & 94 \\ \midrule
    \multicolumn{2}{l}{\textbf{Human Performance}*~\citep{cvpr/alfred}}                         & 91 & - & - & - & -  & -   & -    \\ \bottomrule
    \end{tabular}
    }
    \vspace{-0.5em}
\end{sc}
\end{small}
\end{center}
\end{table*}

\begin{figure}[ht]
\begin{center}
\centerline{\includegraphics[width=0.75\columnwidth]{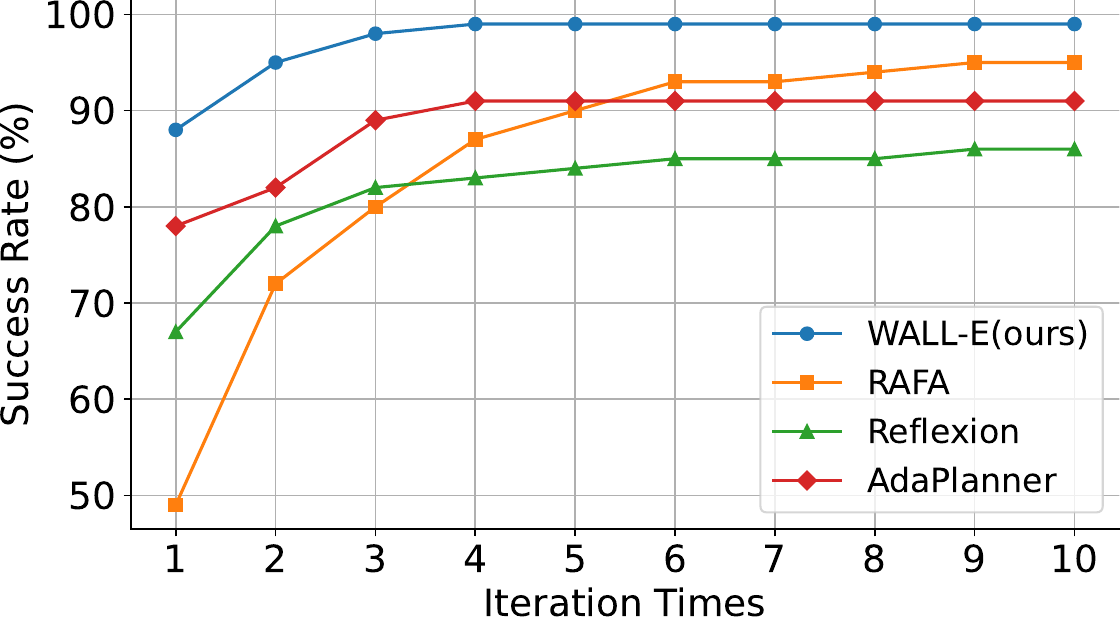}}
\vspace{-1.em}
\caption{\ours vs. baselines on 134 testing tasks from the ALFWorld. \textbf{\ours exhibiting superior planning ability and achieves the highest success rate after only 4 iterations, significantly surpassing other baselines.}}
\label{fig:Alfworld_Success_Rate_Across_Episodes}
\end{center}
\vspace{-2.em}
\end{figure}

\paragraph{\ours excels in planning and task-solving with rapid adaptation.}  
As shown in Tables~\ref{tab:mars-main-result} and \ref{tab:alfworld-main-results}, \ours consistently outperforms baselines across diverse environments. In Mars, \ours achieves reward improvements of 16.1\% to 51.6\% and boosts task scores by at least 61.7\%. While DreamerV3 reaches higher absolute rewards, it is trained for 1 million environment steps—far exceeding \ours’s budget of just 5,600 steps (0.56\% of DreamerV3’s budget). Our focus is not on absolute performance, but on enabling efficient few-shot adaptation and situated reasoning. DreamerV3 is included as a reference to emphasize this contrast in learning efficiency. In ALFWorld, \ours achieves the highest success rate after only four iterations, significantly surpassing strong baselines like RAFA~\citep{hao2023reasoning} and AdaPlanner~\citep{sun2024adaplanner}, as shown in Figure~\ref{fig:Alfworld_Success_Rate_Across_Episodes}.

\paragraph{\ours excels in adapting to environments that contradict the LLM’s prior knowledge, demonstrating strong adaptability.} 
Table~\ref{tab:mars-main-result} shows that when the environment’s mechanics conflict with the LLM’s prior knowledge, the agent’s performance drops substantially, ranging from 3.3\% to 97.1\%, compared to the default setting. Nevertheless, our method can learn symbolic knowledge through inductive reasoning to align its world model with the current environment. This alignment leads to more accurate predictions, feedback, and suggestions during planning, thereby significantly boosting the agent’s performance. In these counter-commonsense scenarios, \ours achieves a reward increase of at least 21.8\% and a score increase of at least 51.6\% compared to other methods.

Furthermore, when environmental changes originate from a single source, our method exhibits strong adaptability, achieving reward and score improvements of at least 31\% and 66\%, respectively.  However, when multiple factors are modified at once, the improvement is less pronounced, likely because our rule representations are tailored to specific dynamic mechanics and cannot fully capture more general, complex mechanics in the environment.

Additionally, in the ``Survival'' setting, our method exhibits a large standard deviation. This arises because in this setting, cows can shoot arrows at the agent: if the initial position spawns a large group of cows, the survival rate plummets, significantly degrading performance. Conversely, if few cows appear near the initial position, our aligned world model secures strong performance. The resulting variation leads to a higher standard deviation for our method.

\begin{figure}[ht]
\vskip 0.2in
\begin{center}
\centerline{\includegraphics[width=0.9\columnwidth]{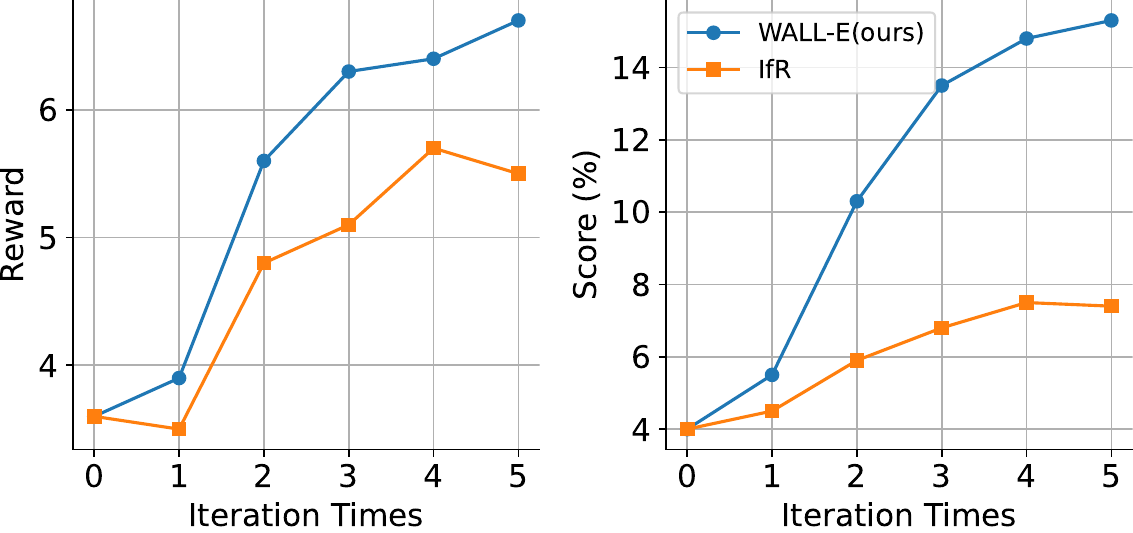}}
\caption{Comparison between \ours and IfR (the best baseline in Table~\ref{tab:mars-main-result}) over learning iterations in Mars. 
\textbf{\ours achieves a clear advantage over IfR in both learning efficiency and overall performance}, due to the world alignment with diverse symbolic knowledge and code rules.
}
\label{fig:Mars-CodeVsNLRules}
\end{center}
\vskip -0.2in
\end{figure}

\paragraph{\ours significantly outperforms the SOTA method IfR.}
Although IfR is the current SOTA on the Mars benchmark, Figure~\ref{fig:Mars-CodeVsNLRules} show that \ours surpasses it in every aspect. As the number of iterations increases, \ours’s reward rises from 3.6 to 6.7, while its score improves from 4\% to 15.3\%, exceeding IfR by 17.9\% and 51.6\%, respectively.
Several factors account for this performance gap. First, although IfR employs inductive reasoning to learn the environment’s mechanisms, it relies solely on rules. In contrast, our NeuroSymbolic learning approach acquires action rules, knowledge graphs, and scene graphs, providing a more comprehensive understanding of the environment and allowing the agent to align more effectively. Additionally, IfR’s rules are expressed in natural language and can only be applied through prompt-based learning, which introduces randomness—particularly when the model’s prior knowledge conflicts with the rules. Our method, however, utilizes code rules, forcing the LLM to adhere to them strictly and thereby reducing variability. This design choice significantly enhances overall performance.

\subsection{Effectiveness of NeuroSymbolic Learning}

\begin{figure}[ht]
\vskip 0.2in
\begin{center}
\centerline{\includegraphics[width=0.9\columnwidth]{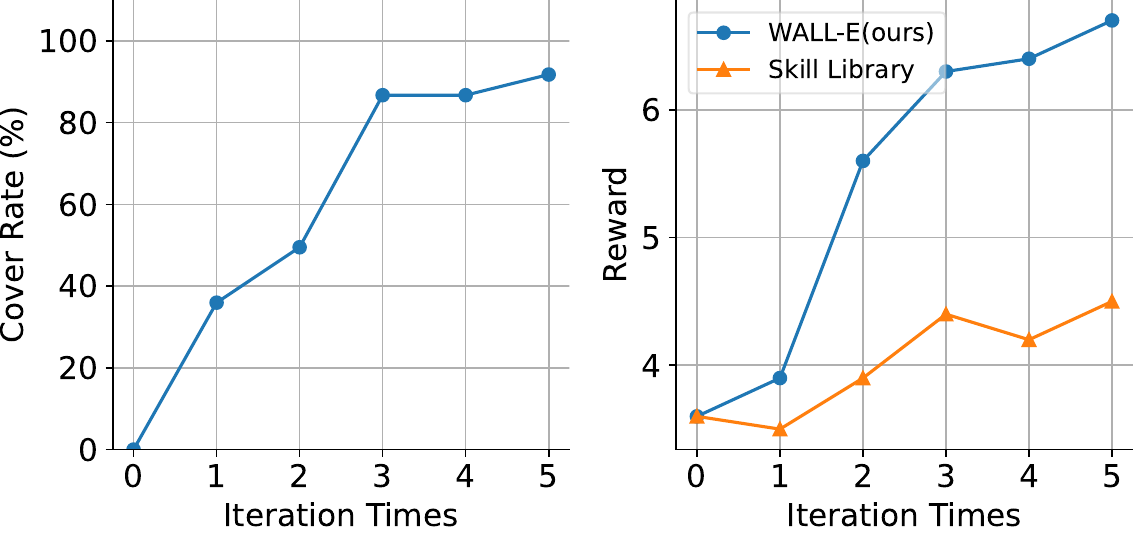}}
\caption{\textbf{Left}: \ours code rules' cover rate (higher the better) over neurosymsbolic learning iterations in Mars. The cover rate measures the percentage of LLM failed predictions that can be corrected by our world model. 
\textbf{Right}: 
Comparison between \ours's world model and skill library when applied to LLM agents with reasoning and reflection (base agent). 
\ours---base agent + world model; 
Skill Library---base agent + skill library.
\textbf{\ours's neurosymbolic learning significantly improves world alignment and brings greater gains to LLM agents than skill library. \looseness-1
}
}
\label{fig:Mars-EffectivenessofRule}
\end{center}
\vskip -0.2in
\end{figure}

To evaluate the effectiveness of our proposed NeuroSymbolic Learning approach, we compare it with the skill library, a widely used agent learning method also employed by JARVIS-1~\citep{wang2024jarvis} and Voyager~\citep{wang2023voyager}. In the Mars benchmark, this approach is further simplified to better adapt to the environment, resulting in a framework consisting of a reasoning module, a reflector, and the skill library~\citep{tang2024mars}. In contrast, \ours comprises a reasoning module, a reflector, and a neurosymbolic learning module. The skill library stores successful plans and utilizes them as in-context examples to align LLM agents with the environment’s dynamics.
By analyzing both the NeuroSymbolic Learning process and the agent’s performance across iterations (Figure~\ref{fig:Mars-EffectivenessofRule}), we observe a significant performance improvement as the quality of the symbolic knowledge library increases (i.e., as the cover rate rises). This result highlights that \ours’s advancements primarily stem from acquiring and leveraging new symbolic knowledge.

\noindent\textbf{NeuroSymbolic Learning enables efficient ``world alignment''.} To verify whether the learned symbolic knowledge lead to a more accurate world model, we collect a dataset of transitions that the LLM world model initially fails to predict. We then calculate the cover rate—the fraction of these mispredictions that are corrected by the newly learned symbolic knowledge (see Appendix ~\ref{sec:Experiment Design for Effectiveness of Rule Learning} for more details). As shown in Figure~\ref{fig:Mars-EffectivenessofRule}, our learned symbolic knowledge progressively raise the cover rate (from 0.0\% to 92.3\%), which in turn boosts performance (reward from 3.6 to 6.7). While the skill library also shows some improvement with additional iterations (reward increasing from 3.6 to 4.5), its gain is less pronounced than \ours’s; within five iterations, \ours’s reward improves by 32.8\% more than the skill library. \looseness-1

\subsection{Ablation Study}

\paragraph{Combining multiple types of symbolic knowledge enhances world alignment }
To evaluate how different types of symbolic knowledge affect world alignment, we compare methods that combine various forms of symbolic knowledge across different environments. Whenever action rules are included, they are implemented as code rules; otherwise, if only knowledge graphs or scene graphs are used, the system relies on prompt learning. As shown in Table~\ref{tab:ablation_study_knowledge}, using only rules or only KG/SG results in a noticeable drop in performance compared with using both. In the more complex Mars environment, using only rules reduces reward and score by 23.9\% and 45.8\%, respectively, while using only KG/SG reduces them by 34.3\% and 66.0\%. These findings underscore the necessity of leveraging multiple types of symbolic knowledge to effectively align the LLM’s world model with the environment.

\begin{table}[t]
\caption{Ablation study of \ours with different symbolic knowledge (KNWL) types.
KG---Knowledge Graph, SG---Scene Graph.
The row highlighted in grey represents the configuration and performance of \ours.
\textbf{Learning diverse types of symbolic knowledge is essential for effectively aligning the LLM world model with the environment.}
}
\label{tab:ablation_study_knowledge}
\begin{center}
\begin{small}
\begin{sc}
\vspace{-1.5em}
\renewcommand\arraystretch{0.9}
\resizebox{\linewidth}{!}{
    \begin{tabular}{lcccc}
    \toprule
    \multicolumn{2}{l}{\textbf{Symbolic KNWL}} & \multicolumn{2}{c}{\textbf{Mars}}                           & \textbf{ALFWorld}         \\  \cmidrule(lr){1-2}   \cmidrule(lr){3-4}  \cmidrule(lr){5-5} 
     Action Rules          &     KG/SG             & \textbf{Reward $\uparrow$} & \textbf{Score (\%) $\uparrow$} & \textbf{Success Rate (\%) $\uparrow$} \\ \midrule
     \checkmark     &                       &           5.1              &            8.3                 &         95                \\  
                    &     \checkmark        &            4.4             &           5.2                  &         88                \\  
    \rowcolor{gray!30}  \checkmark  &  \checkmark &      6.7             &           15.3                 &         98                \\    \bottomrule
    \end{tabular}
}
\vspace{-1em}
\end{sc}
\end{small}
\end{center}
\end{table}

\paragraph{World models with symbolic knowledge is an effective structure.}
We examine the impact of learned symbolic knowledge and the world model by removing each component from \ours and observing changes in performance (Table~\ref{tab:ablation_study_structure}). Adding symbolic knowledge, whether in the agent or the world model, consistently boosts the success rate. Specifically, applying symbolic knowledge in the world model yields about a 46.3\% improvement, while applying them in the agent yields about a 30.9\% gain. This difference likely arises because learned symbolic knowledge heavily depend on observation information (see Appendix~\ref{sec:Learned Rules}). Moreover, a standalone world model without symbolic knowledge offers little performance benefit, underscoring that symbolic knowledge-driven alignment with environment mechanism is key to \ours’s success.

\begin{table}[t]
\caption{Ablation study of \ours with different configurations on Mars tasks. Symbolic---apply code rules in the prompt.  
The row highlighted in grey represents the configuration and performance of \ours.
\textbf{Code rules translated from symbolic knowledge bring more improvement when applied to the world model, indicating the importance of world alignment.  
}}
\label{tab:ablation_study_structure}
\begin{center}
\begin{small}
\begin{sc}
\vspace{-1em}
\renewcommand\arraystretch{0.85}
\resizebox{\linewidth}{!}{
    \begin{tabular}{llcc}
    \toprule
    Agent         & World Model    & Reward $\uparrow$      & Score (\%) $\uparrow$   \\ 
    \cmidrule(lr){1-2} \cmidrule(lr){3-4} 
    LLM                    & -              &  3.6     &   4.0    \\ 
    LLM                    & LLM            &  3.8      &  4.1      \\
    LLM+symbolic              & -              &  5.5     &   7.4     \\ 
    \rowcolor{gray!30} LLM & LLM+symbolic      &  6.7      &   15.3      \\ 
    LLM+symbolic              & LLM+symbolic      &  6.3      &   13.1       \\ 
    \bottomrule
    \end{tabular}
}
\end{sc}
\end{small}
\end{center}
\end{table}

\paragraph{Ablation on code rule pruning stage.}
We conducted ablation studies on the code rule pruning stage, including the performance drop when skipping it and how varying the code rule set limit $l$ affects outcomes (Table~\ref{tab:ablation_rule_pruning_stage}). The table shows that pruning stage is essential—without it, performance drops sharply due to noisy or conflicting rules. As fewer rules are retained, performance degrades, confirming that pruning selects impactful, high-quality rules that align the LLM with environment dynamics.

\begin{table}[h]
\centering
\caption{Ablation study on the code rule set pruning stage in Mars. The column
highlighted in \textbf{bold} represents the configuration and performance
of \ours. \textbf{Pruning is essential for selecting high-quality, impactful rules that align the LLM world model with environment dynamics.}}
\label{tab:ablation_rule_pruning_stage}
\resizebox{\linewidth}{!}{
\begin{tabular}{lccccc}
\toprule
\multirow{2}{*}{Metrics} & \multirow{2}{*}{w/o Code Rule Set Pruning} & \multicolumn{4}{c}{Code Rule Set Limit} \\ \cmidrule(lr){3-6}
 &  & \textbf{no limit (6)} & 5 & 3 & 1 \\
\midrule
Reward $\uparrow$ & 1.5 & \textbf{6.7} & 6.2 & 5.7 & 4.3 \\
Score (\%) $\uparrow$ & 1.6 & \textbf{15.3} & 12.8 & 9.5 & 8.1 \\
\bottomrule
\end{tabular}
}
\end{table}

\section{Conclusion}

We have demonstrated that LLMs can effectively serve as world models for agents when aligned with environment dynamics via neurosymbolic knowledge learning. Our neurosymbolic approach leverages code-based, gradient-free integrations of action rules, knowledge graphs, and scene graphs, thereby bridging the gap between the LLMs' prior knowledge and specific environments. By using a model-based framework, our agent, \ours, achieves substantial improvements in planning and task completion. In open-world environments such as Mars and ALFWorld, \ours outperforms baselines by 16.1\%–51.6\% in success rate and achieves a new record 98\% success rate in ALFWorld after only four iterations. 
These results underscore that additional symbolic knowledge is essential to align LLM predictions with environment dynamics, enabling high-performing model-based agents in complex settings.

\bibliography{walle2}
\bibliographystyle{walle2}

\newpage
\onecolumn
\appendix

\section{Detailed Related Work} 
\label{sec:Detailed Related Work}

\paragraph{LLMs with Rule Learning.}
Recent studies have explored integrating LLMs with rule learning to enhance reasoning and model behavior. For instance, \cite{yang2023enabling} introduced rule distillation, enabling LLMs to learn from predefined rules, which improved generalization with limited training data. Similarly, \cite{Zhu2023Large} proposed the Hypotheses-to-Theories (HtT) framework, which enhanced numerical and relational reasoning by generating and validating rules from training data. In the same vein, \cite{mu2023can} developed the RuLES framework to evaluate LLM adherence to developer-specified rules, addressing challenges like rule evasion through adversarial inputs. Furthermore, \cite{yang2023failures} presented the Tuning-free Rule Accumulation (TRAN) framework, allowing LLMs to accumulate rules from incorrect cases to avoid repeating mistakes without additional tuning. Lastly, in knowledge graph reasoning, \cite{luo2023chatrule} introduced ChatRule, a framework that mines logical rules over knowledge graphs using LLMs.

These studies show the potential of combining LLMs with rule learning to improve reasoning and generalization. However, none have integrated rule learning with LLM-based world models, which is the focus of our work. We explore how rule learning can align LLM world models with specific environment dynamics, thereby improving the performance of model-based agents in dynamic environments.

\paragraph{Using LLMs to Build World Models.}
Many studies have leveraged LLMs to construct world models for planning. For example, \cite{wong2023learning} proposed translating natural language instructions into adaptable planning representations via LLMs, enabling flexible and context-aware world modeling. Similarly, \cite{guan2023leveraging} showed that combining pre-trained LLMs with task-specific planning modules improves task success rates by providing a more detailed understanding of the environment. Another approach, WorldCoder \cite{tang2024worldcoder}, exemplified an LLM agent that constructs world models by generating and executing code to simulate various states and actions, refining its understanding iteratively.

These studies demonstrate the effectiveness of using LLMs to construct explicit world models from scratch to enhance planning and reasoning in complex environments. However, our key novelty lies in treating the LLM itself as the world model and aligning it with environment dynamics through training-free symbolic knowledge extraction. Rather than building a complete symbolic model from the ground up, we focus on identifying and correcting the misalignment between the base LLM’s prior knowledge and the specific environment. This allows a small amount of complementary symbolic knowledge, combined with the LLM’s existing priors, to form an accurate and generalizable world model. In contrast, prior methods rely on complex, environment-specific modeling procedures, which limits their scalability and adaptability to high-complexity environments like Mars.

\paragraph{Using LLMs as World Models.}
Several studies have explored using LLMs directly as world models by leveraging their implicit knowledge. Some methods rely on fine-tuning to align the LLM world model with the environment. For example, \cite{xiang2024language} fine-tuned LLMs with embodied experiences in a simulated world to enhance reasoning and planning abilities in embodied environments. Similarly, \cite{xie2024making} transformed LLMs into world models by incorporating knowledge of action preconditions and effects, fine-tuning the models to reason about actions and predict their outcomes accurately.

Other approaches align LLMs as world models through prompting. For instance, \cite{zhao2024large} introduced the LLM-MCTS algorithm, prompting LLMs to serve as both the policy and world model for large-scale task planning, integrating commonsense priors with guided search. In another approach, \cite{hao2023reasoning} introduced Reasoning via Planning (RAP), where LLMs are prompted to act as reasoning agents and world models by generating reasoning trees to explore solutions. Finally, \citep{liu2023reason} used a Bayesian adaptive Markov Decision Process to guide LLMs in planning future trajectories, prompting them to predict future states.

While these approaches demonstrate the potential of using LLMs as world models, they often require extensive fine-tuning or rely heavily on human-crafted prompts, making them labor-intensive and inflexible. Our work overcomes these limitations by automatically extracting rules from exploration experiences, reducing human effort and enhancing adaptability across different environments.

\subsection{Existing Agents in Minecraft} 
\label{sec:Existing Agents in Minecraft}
Although our method is evaluated in Mars, a 2D environment inspired by Minecraft, we also discuss prior agent methods developed in the Minecraft domain due to their conceptual relevance.

DEPS \cite{wang2023describe} uses an LLM to generate, explain, and revise plans, but it does NOT learn knowledge from previous tasks and generalize to new scenarios. 
Methods like GITM \cite{zhu2023ghost}, MP5 \cite{qin2024mp5}, JARVIS-1 \cite{wang2024jarvis}, and Voyager \cite{wang2023voyager} use a memory buffer or skill library to directly store successful trajectories and reuse them later. However, no inductive reasoning is applied to these trajectories to learn condensed, generalized principles. So their agents have to reason based on many raw trajectories, which can suffer from poor generalization, overfitting, and inefficiency when adapted to new scenarios/tasks.
Optimus-1 \cite{li2024optimus} partially addresses this issue by extracting knowledge graphs from historical trajectories, enabling more flexible knowledge transfer across tasks. However, it still relies heavily on logging trajectories and custom scripts tailored to a specific environment's data format for graph extraction, which cannot be easily generalized to new environments.

In contrast, our method does not store and reason directly based on raw trajectories. It overcomes these limitations by inductive reasoning using LLM on existing trajectories, condensing them into a compact, principal, generalizable, and comprehensive set of symbolic knowledge (action rules, knowledge graphs, and scene graphs) that can be reused efficiently and generalized across tasks/environments. This LLM-abstraction of environment is novel and critical to building an accurate world model. Another primary difference and novelty of \ours is to compile all symbolic knowledge into executable code, which makes the planning verifiable with formal guarantees. This avoids possible hallucinations and mistakes of in-context learning by LLMs, which are common for planning in prior work.

\subsection{Existing Agents in ALFWorld} 
\label{sec:Existing Agents in ALFWorld}

ReAct \cite{yao2023react}, Reflexion \cite{shinn2024reflexion}, and AutoGen \cite{wu2023autogen} leverage LLMs for reasoning, but they do NOT learn reusable knowledge from past tasks. 
AdaPlanner \cite{sun2024adaplanner} introduces a skill library, but it merely stores trajectories from previous successful tasks. Reasoning solely based on these specific trajectories can mislead the agent and lead to suboptimal behavior in new scenarios.
RAFA \cite{liu2023reason} relies on an LLM’s prior knowledge for planning, which can be problematic when the environment dynamics and principles diverge from or conflict with LLM priors. 

\ours differs from them in several key properties: Instead of simply recording past trajectories, \ours prompts the LLM to perform inductive reasoning over historical trajectories to extract compact and generalizable symbolic knowledge, as discussed above. Unlike RAFA, whose model LLM solely relies on LLM prior, \ours applies verifiable neurosymbolic learning to align the LLM’s predictions with real environment dynamics. This ``world alignment'' greatly enhances both prediction accuracy and long-horizon planning capabilities.

\newpage
\section{Detailed Prompt}

\subsection{Learn Action Rules from Real Trajectories}
\label{sec:Prompt for Learn New Rules from Real Trajectories}
\begin{center}
    \textbf{Prompt for Learning Action Rules from Real Trajectories}
\end{center}
\begin{adjustwidth}{\dimexpr(\textwidth-0.95\textwidth)/2}{}
\lstset{language=}
\begin{lstlisting}[linewidth=0.95\textwidth]
You are responsible for mining new rules from the given transitions, ensuring that 
these rules differ from the ones already provided. Focus on generating general and 
universal rules that are not tied to any specific item or tool. Your goal is to 
generalize across different objects, creating flexible rules that can be applied 
broadly to diverse contexts and situations.

I will give you an array of transitions:
[
    {
        'state_0': {
            "state feature 1": {"feature name": value, ...},
            ...
        }, 
        'action': {
            "name": "action name", 
            "action feature 1": {"feature name": value, ...},
            ...
        }, 
        'action_result': {
        "feedback": "the environment feedback",
        "success": "Whether the action is executed successfully,",
        "suggestion": "..."
    }
    },
    ...
]
and an array of rules:
[
    "Rule 1: For action ..., if..., the action will fail; Checking Method: ...",
    ...
]

You should only respond in the format as described below:
RESPONSE FORMAT:
{
    "new_rules":[
        "Rule ...: For action ...,...; Checking Method: ...",
        "Rule ...: For action ...,...; Checking Method: ...",
        ...
    ]
}

Instructions:
- Ensure the response can be parsed by Python `json.loads`, e.g.: no trailing 
commas, **no single quotes**, etc.
- Please use you knowledge in <ENV>, do inductive reasoning. You need to dig up 
as many rules as possible that satisfy all transitions.
- Extract and utilize only the features that influence the outcome of the action.
- Please generate general and universal rules; the rules should not reference 
any specific item or tool! You need to generalize across various items or tools.
- Generate only the rules under what conditions the action will fail.
- While generating a rule, you also need to state how to check if a transition 
satisfies this rule. Please be specific as to which and how 'features' need to 
be checked
\end{lstlisting}
\end{adjustwidth}

\newpage
\subsection{Translate Action Rules to Code }
\label{sec:Translate Natural Language Rules to Code}

\begin{center}
    \textbf{Prompt for Translating Action Rules to Code}
\end{center}

\begin{adjustwidth}{\dimexpr(\textwidth-0.95\textwidth)/2}{}
\lstset{language=}
\begin{lstlisting}[linewidth=0.95\textwidth]
You are responsible for generating code rules by implementing the learned rules 
in Python. Your task is to write a function that takes the current state and an 
action as inputs, optionally incorporating both the knowledge graph and the 
scene graph to provide additional contextual information (extra input), evaluating 
these conditions, and returns a Boolean value based on the specified rule. 
This function should effectively mirror the logic of the rules, enabling precise 
predictions for various state-action pairs.

The function should be defined as follows:

```python
def expected_rule_code(state, action, knowledge_graph/scene_graph):
    # Your code here
    return feedback, success, suggestion
where
feedback: a string, give the action feedback based on success or not.
success: a bool, whether the action is executed successfully, give 'True' or 
'False'. If the action type is not the action type in the rule, count as success 
(e.g., success = True).
suggestion: a string, if the 'action' fails, 'suggestion' would be given based on 
'rule', 'state' and 'action'.

Here are examples of the state and action format:
<Input Format>

For example:
    "Instead of obtaining [item] from [collecting_resource], players can acquire 
    it from [alternative collecting_resource].",
    "To craft a [tool/object], players will need [crafting_material] and must 
    use [crafting_platform].",
    "[crafting_material] needs to be gathered from [resource].",
    ...
Be sure to include such details to make the suggestions more engaging and relevant.

Knowledge Graph:
<Knowledge Graph>

Scene Graph:
<Scene Graph>

You should only respond in the format as described below, and do not give example 
usage or anything else:
RESPONSE FORMAT:
def expected_rule_code(state, action, knowledge_graph/scene_graph):
    # Your code here
\end{lstlisting}
\end{adjustwidth}
where ``input format'' please refer to Appendix \ref{sec: Environments' State Space and Action Space}.

\newpage
\subsection{Learn Knowledge Graph from Real Trajectories}
\label{sec:Prompt for Learn Knowledge Graph from Real Trajectories}

\begin{center}
    \textbf{Prompt for Learning Knowledge Graph from Real Trajectories}
\end{center}

\begin{adjustwidth}{\dimexpr(\textwidth-0.95\textwidth)/2}{}
\lstset{language=}
\begin{lstlisting}[linewidth=0.95\textwidth]
You are a helpful assistant with inductive reasoning. Given the history trajectory, 
including action and observation, you need to reflect on the action execution results 
and identify and extract prerequisite or feasibility constraints, that is, discover 
when an action or item creation requires the presence of certain materials, resources, or other items.

We define the Knowledge Graph as:
{
  "V": "the set of entities (e.g., items, materials, location-specific objects, or 
  abstract concepts)",
  "E": "the set of directed edges, each capturing a relationship or prerequisite 
  among entities"
}

An edge takes the form:
(u, v, label),
where u and v are entities in V, and label indicates how u relates to v 
(for example, 'requires', 'consumes', 'collects', etc.).

I will give you an array of transitions:
[
    {
        'inital_state': '...', 
        'action': '...', 
        'action_result': "Whether the action is executed successfully, give 'True' 
        or 'False' only"
    },
    {
        'inital_state': '...', 
        'action': '...', 
        'action_result': "Whether the action is executed successfully, give 'True' 
        or 'False' only"
    },
    ...
]

You should ONLY respond in the following format:
{
    {'u':'entity_u', 'v':'entity_v', 'label':{'relation':'...', 'quantity':'...'}},
    {'u':'entity_u', 'v':'entity_v', 'label':{'relation':'...', 'quantity':'...'}},
    ...
}
example:
{'u':'wooden_sword', 'v':'table', 'label':{'relation':'requires', 'quantity':None}},
{'u':'table', 'v':'wood', 'label':{'relation':'consumes', 'quantity':'2'}}
\end{lstlisting}
\end{adjustwidth}

\newpage
\section{Environments' Observation Space and Action Space}
\label{sec: Environments' State Space and Action Space}

The format of observation and action information is crucial for understanding the action rules we have extracted. In this section, we provide an description of the observation and action space used in different environments.

\subsection{Mars}

\textbf{Observation Space.} We collect observation information directly from the observation space provided by Mars~\citep{tang2024mars}. The specific structure is illustrated in the following example.

\begin{center}
    \textbf{Examples for Mars's Observation Space}
\end{center}
\begin{adjustwidth}{\dimexpr(\textwidth-0.95\textwidth)/2}{}
\lstset{language=}
\begin{lstlisting}[linewidth=0.95\textwidth]
obs = {
    "position": "grass",
    "in_front": "table",
    "visible_objects": [
        {
            "type": "plant",
            "x": -1,
            "y": 0
        },
        {
            "type": "grass",
            "x": 1,
            "y": 0
        },
        {
            "type": "table",
            "x": 0,
            "y": -1
        },...
    ],
    "near_objects": [
        "sand",
        "plant",
        "grass",
        "table"
    ],
    "status": {
        "health": 8,
        "food": 4,
        "drink": 5,
        "energy": 8
    },
    "inventory": {
        "sapling": 2,
        "wood_pickaxe": 2,
        "wood_sword": 1
    }
}
\end{lstlisting}
\end{adjustwidth}

\textbf{Action Space.} We utilize the action space provided by the Mars directly, as demonstrated below. 

\begin{center}
    \textbf{Mars's Action Space}
\end{center}
\begin{adjustwidth}{\dimexpr(\textwidth-0.95\textwidth)/2}{}
\lstset{language=}
\begin{lstlisting}[linewidth=0.95\textwidth]
mine(block_name, amount) # mine amount blocks of the block_name. 
attack(creature, amount) # attack the amount creature that can move. Creature include zombie, skeleton, cow, etc.
sleep(); # put the player to sleep.
place(block_name); # place the block. Note you need not craft table and furnace, you can place them directly.
make(tool_name); # craft a tool.
explore(direction, steps); # the player explore in the direction for steps.
\end{lstlisting}
\end{adjustwidth}

\subsection{ALFWorld}
\label{sec: ALFworld Environments' State Space and Action Space}

\textbf{Observation Space.} In the original ALFWorld setup, observation information is represented as natural language dialogue history. To facilitate the rule learning process, we developed scripts to transform this dialogue history into a structured JSON format, as shown in the following example.

\begin{center}
    \textbf{Examples for ALFWorld's Observation Space}
\end{center}
\begin{adjustwidth}{\dimexpr(\textwidth-0.95\textwidth)/2}{}
\lstset{language=}
\begin{lstlisting}[linewidth=0.95\textwidth]
obs = {
    "reachable_locations": [
        "cabinet 5",
        "cabinet 4",
        "cabinet 3",
        "cabinet 2",
        "cabinet 1",
        "coffeemachine 1",
        "countertop 2",
        "countertop 1",
        "diningtable 1",
        "drawer 2",
        "drawer 1",
        "fridge 1",
        "garbagecan 1",
        "microwave 1",
        "shelf 3",
        "shelf 2",
        "shelf 1",
        "sinkbasin 1",
        "stoveburner 4",
        "stoveburner 3",
        "stoveburner 2",
        "stoveburner 1",
        "toaster 1"
    ],
    "items_in_locations": {
        "fridge 1": [
            "lettuce 2",
            "mug 2",
            "potato 3"
        ],
        "microwave 1": []
    },
    "item_in_hand": {
        "item_name": "cup 1",
        "status": "normal"
    },
    "current_position": {
        "location_name": "microwave 1",
        "status": "open"
    }
}
\end{lstlisting}
\end{adjustwidth}

\newpage
\textbf{Action Space.} We utilize the action space provided by the ALFWorld directly, as demonstrated below. 
\begin{center}
    \textbf{Action Space for ALFWorld}
\end{center}
\begin{adjustwidth}{\dimexpr(\textwidth-0.95\textwidth)/2}{}
\lstset{language=}
\begin{lstlisting}[linewidth=0.95\textwidth]
go to [location/object]: Move to a specified location or object. 
open [object]: Open a specified object like a cabinet or drawer. 
close [object]: Close an opened object.
take [object] from [location]: Pick up an item from a specified location.
put [object] in/on [location]: Place an item in or on a specified location.
clean [object] with [location/tool]: Clean an object using a specific location or tool, like cleaning lettuce at the sink basin.
heat [object] with [tool]: Use an appliance, such as a microwave, to heat an item.
cool [object] with [tool]: Use a cooling tool or appliance, such as a fridge, to cool an item.
use [tool]: Activate or use a tool, such as a desklamp.
\end{lstlisting}
\end{adjustwidth}

\newpage
\section{Learned Rules}
\label{sec:Learned Rules}

There are two points to note about the numbering of the rules:
\begin{itemize}
    \item The reason for duplicates is that the numbering is based on actions, and different actions have their own separate sequences. For example: Rules for Craft: [Rule 1, Rule 2, Rule 3, Rule 4, Rule 5...]; Rules for Mine: [Rule 1, Rule 2, Rule 3, Rule 4, Rule 5...].
    \item The reason the sequence may appear unordered is that some rules have been pruned (Section \ref{sec:Translating Symbolic Knowledge  to Code} Rule Set Pruning via Maximum Coverage). For instance, Rules for Craft where [Rule 1, Rule 2, Rule 4, Rule 5] has been removed, Rules for Mine where [Rule 1, Rule 3, Rule 4, Rule 5, Rule 6] has been removed, and the final rule set is Rules for Craft: [Rule 3, Rule 6] and Rules for Mine: [Rule 2, Rule 7].
\end{itemize}

Additionally, the feedback and suggestions returned by each code rule are automatically generated by prompting the LLM with the corresponding rule. The detailed prompts used to generate these code rules can be found in Appendix \ref{sec:Translate Natural Language Rules to Code}. These feedback and suggestions play a crucial role in helping the agent refine and improve its planning process (Section \ref{sec:MPCinWALLE}).

\subsection{Rules in Mars}

\begin{center}
    \textbf{Action Rules for Mars}
\end{center}
\begin{adjustwidth}{\dimexpr(\textwidth-0.95\textwidth)/2}{}
\lstset{language=}
\begin{lstlisting}[linewidth=0.95\textwidth]
Rule 6: For action 'make', if 'table' is not in 'near_objects', the action will fail; 
Checking Method: Check if 'table' is in the 'near_objects' list of the initial state. 
Rule 7: For action 'make', if the player does not have the required materials in the 
inventory to craft the specified tool, the action will fail; Checking Method: Verify 
if the player's inventory contains the required materials for the tool specified in 
the 'make' action's 'tool_name' argument. 
Rule 8: For action 'make', if the required resources for the tool are not present in 
the inventory or if a 'table' is not in 'near_objects', the action will fail; 
Checking Method: Verify if the inventory contains the necessary resources for the 
tool being 
crafted and ensure that 'table' is present in 'near_objects'. 
Rule 1: For action 'place' with block_name 'sapling', if 'in_front' is 'table', the 
action will fail; Checking Method: Check if 'in_front' in the initial state is 
'table'. 
Rule 2: For action 'place', if the player does not have the required materials in 
the inventory to place the specified item, the action will fail; Checking Method: 
Verify if the player's inventory contains the required materials for the item 
specified in the 'place' action's 'block_name' argument. 
Rule 3: For action 'place' with block_name 'table', if 'visible_objects' or 
'near_objects' already contain a 'table', the action should not be executed; 
Checking Method: Examine both the 'visible_objects' and 'near_objects' arrays in the 
state for an object with type 'table'. If found, do not proceed with placing a new 
table, and instead reuse the existing one. 
Rule 2: For action 'mine', if the block_name is 'iron' and the player does not have 
'stone_pickaxe' or better, the action will fail; Checking Method: Verify if the 
'block_name' is 'iron' and ensure the inventory includes a 'stone_pickaxe' or better. 
Rule 4: For action 'mine', if the player attempts to mine 'stone' and does not have 
a 'wood_pickaxe' or better in their inventory, the action will fail; Checking Method: 
Check if 'wood_pickaxe' or a better tool is present in the inventory when attempting 
to mine 'stone'. 
Rule 5: For action 'mine', if the block_name is 'plant', the action will fail; 
Checking Method: Check if 'block_name' in the action is 'plant'. 
\end{lstlisting}
\end{adjustwidth}

\newpage
\begin{center}
    \textbf{Code Rules for Mars}
\end{center}
\begin{adjustwidth}{\dimexpr(\textwidth-0.95\textwidth)/2}{}
\lstset{language=python}
\begin{lstlisting}[linewidth=0.95\textwidth]
def Rule_6_make(state, action, knowledge_graph):
    # Extract action details
    action_name = action.get("action_name")
    args = action.get("args", {})
    
    # Check if the action is 'make'
    if action_name == "make":
        # Check if 'table' is in 'near_objects'
        if "table" not in state.get("near_objects", []):
            feedback = "Action failed: 'table' is not nearby."
            success = False
            suggestion = "Move closer to a 'table' to make the item."
            return feedback, success, suggestion
    
    # If the action is not 'make' or the condition is met
    feedback = "Action executed successfully."
    success = True
    suggestion = ""
    return feedback, success, suggestion
\end{lstlisting}
\end{adjustwidth}
\begin{adjustwidth}{\dimexpr(\textwidth-0.95\textwidth)/2}{}
\lstset{language=python}
\begin{lstlisting}[linewidth=0.95\textwidth]
def Rule_7_make(state, action, knowledge_graph):
    # Check if the action is 'make'
    if action['action_name'] == 'make':
        tool_name = action['args']['tool_name']
        
        # Use LLM_request to find out the required materials for the tool
        question = f"What are the required materials to craft a {tool_name}?"
        response_format = "Provide a list of materials and their quantities."
        required_materials = LLM_request(question + response_format)
        
        # Check if the player has the required materials in the inventory
        inventory = state['inventory']
        has_all_materials = True
        missing_materials = []
        
        for material, quantity in required_materials.items():
            if inventory.get(material, 0) < quantity:
                has_all_materials = False
                missing_materials.append(f"{material}: {quantity - 
                inventory.get(material, 0)} more needed")
        
        if has_all_materials:
            feedback = "Action executed successfully."
            success = True
            suggestion = ""
        else:
            feedback = "Action failed: Not enough materials to craft the tool."
            success = False
            suggestion = f"To craft a {tool_name}, you need: 
            {', '.join(missing_materials)}."
    else:
        # If the action is not 'make', consider it successful
        feedback = "Action executed successfully."
        success = True
        suggestion = ""
    
    return feedback, success, suggestion
\end{lstlisting}
\end{adjustwidth}
\newpage
\begin{adjustwidth}{\dimexpr(\textwidth-0.95\textwidth)/2}{}
\lstset{language=python}
\begin{lstlisting}[linewidth=0.95\textwidth]
def Rule_8_make(state, action, knowledge_graph):
    # Extract action details
    action_name = action.get("action_name")
    args = action.get("args", {})
    
    # Check if the action is 'make'
    if action_name != "make":
        return "Action executed successfully.", True, ""
    
    # Extract tool name
    tool_name = args.get("tool_name")
    
    # Check if 'table' is in 'near_objects'
    if 'table' not in state.get("near_objects", []):
        feedback = "Action failed: A 'table' is required nearby to make the tool."
        suggestion = "Move closer to a 'table' and try again."
        return feedback, False, suggestion
    
    # Check if the required resources are in the inventory
    required_resources = knowledge_graph.get(tool_name, {})
    inventory = state.get("inventory", {})
    
    for resource, amount in required_resources.items():
        if inventory.get(resource, 0) < amount:
            feedback = f"Action failed: Not enough {resource} to make {tool_name}."
            suggestion = f"Collect more {resource} to make {tool_name}."
            return feedback, False, suggestion
    
    # If all checks pass
    return "Action executed successfully.", True, ""
\end{lstlisting}
\end{adjustwidth}
\begin{adjustwidth}{\dimexpr(\textwidth-0.95\textwidth)/2}{}
\lstset{language=python}
\begin{lstlisting}[linewidth=0.95\textwidth]
def Rule_1_place(state, action, knowledge_graph):
    # Extract action details
    action_name = action.get("action_name")
    block_name = action.get("args", {}).get("block_name")

    # Initialize feedback, success, and suggestion
    feedback = ""
    success = True
    suggestion = ""

    # Rule 1: For action 'place' with block_name 'sapling', if 'in_front' is 'table', 
    the action will fail
    if action_name == "place" and block_name == "sapling":
        if state.get("in_front") == "table":
            feedback = "Action failed: Cannot place a sapling in front of a table."
            success = False
            suggestion = "Try placing the sapling in front of a different object, such 
            as grass or dirt."
        else:
            feedback = "Action succeeded: Sapling placed successfully."

    # If the action type is not 'place', count as success
    else:
        feedback = "Action succeeded: No rules apply to this action type."

    return feedback, success, suggestion

\end{lstlisting}
\end{adjustwidth}
\newpage
\begin{adjustwidth}{\dimexpr(\textwidth-0.95\textwidth)/2}{}
\lstset{language=python}
\begin{lstlisting}[linewidth=0.95\textwidth]

def Rule_2_place(state, action, knowledge_graph):
    # Check if the action is 'place'
    if action['action_name'] == 'place':
        block_name = action['args']['block_name']
        
        # Check if the required material is in the inventory
        if block_name in state['inventory'] and state['inventory'][block_name] > 0:
            feedback = f"Successfully placed {block_name}."
            success = True
            suggestion = ""
        else:
            feedback = f"Failed to place {block_name}."
            success = False
            suggestion = f"You need to have {block_name} in your inventory to place it. 
            Consider collecting or crafting it first."
    else:
        # If the action is not 'place', consider it successful
        feedback = "Action executed successfully."
        success = True
        suggestion = ""

    return feedback, success, suggestion
\end{lstlisting}
\end{adjustwidth}
\begin{adjustwidth}{\dimexpr(\textwidth-0.95\textwidth)/2}{}
\lstset{language=python}
\begin{lstlisting}[linewidth=0.95\textwidth]
def Rule_3_place(state, action, knowledge_graph):
    # Extract action details
    action_name = action.get("action_name")
    block_name = action.get("args", {}).get("block_name")

    # Check if the action is 'place' and block_name is 'table'
    if action_name == "place" and block_name == "table":
        # Check for existing 'table' in 'visible_objects' or 'near_objects'
        visible_objects = state.get("visible_objects", [])
        near_objects = state.get("near_objects", [])

        # Check if 'table' is in visible_objects
        table_in_visible = any(obj.get("type") == "table" for obj in visible_objects)
        # Check if 'table' is in near_objects
        table_in_near = "table" in near_objects

        # If a table is found in either, the action should not be executed
        if table_in_visible or table_in_near:
            feedback = "Action failed: A table is already present in the vicinity."
            success = False
            suggestion = "Reuse the existing table instead of placing a new one."
            return feedback, success, suggestion

    # If the action is not 'place' or block_name is not 'table', consider it successful
    feedback = "Action executed successfully."
    success = True
    suggestion = ""
    return feedback, success, suggestion
\end{lstlisting}
\end{adjustwidth}
\newpage
\begin{adjustwidth}{\dimexpr(\textwidth-0.95\textwidth)/2}{}
\lstset{language=python}
\begin{lstlisting}[linewidth=0.95\textwidth]
def Rule_2_mine(state, action, knowledge_graph):
    # Extract action details
    action_name = action.get("action_name")
    block_name = action.get("args", {}).get("block_name")
    
    # Check if the action is 'mine' and the block_name is 'iron'
    if action_name == "mine" and block_name == "iron":
        # Check if the player has a 'stone_pickaxe' or better in the inventory
        inventory = state.get("inventory", {})
        has_stone_pickaxe_or_better = any(
            tool in inventory for tool in ["stone_pickaxe", "iron_pickaxe", 
            "diamond_pickaxe"]
        )
        
        if not has_stone_pickaxe_or_better:
            feedback = "Action failed: You need a stone_pickaxe or better to mine iron."
            success = False
            suggestion = "Consider crafting or acquiring a stone_pickaxe or better to 
            mine iron."
            return feedback, success, suggestion
    
    # If the action is not 'mine' or the block_name is not 'iron', consider it successful
    feedback = "Action executed successfully."
    success = True
    suggestion = ""
    return feedback, success, suggestion
\end{lstlisting}
\end{adjustwidth}
\begin{adjustwidth}{\dimexpr(\textwidth-0.95\textwidth)/2}{}
\lstset{language=python}
\begin{lstlisting}[linewidth=0.95\textwidth]
def Rule_4_mine(state, action, knowledge_graph):
    # Extract action details
    action_name = action.get("action_name")
    block_name = action.get("args", {}).get("block_name")
    # Check if the action is 'mine' and the block is 'stone'
    if action_name == "mine" and block_name == "stone":
        # Check if 'wood_pickaxe' or a better tool is in the inventory
        inventory = state.get("inventory", {})
        has_wood_pickaxe = 'wood_pickaxe' in inventory
        better_tool_exists = False
        for tool in inventory:
            if tool != 'wood_pickaxe':
                response = LLM_request(f"is {tool} better than wood_pickaxe?" + 
                "only reply True or False")
                if response == "True":
                    better_tool_exists = True
                    break
        if has_wood_pickaxe or better_tool_exists:
            feedback = "Action executed successfully."
            success = True
            suggestion = ""
        else:
            feedback = "Action failed: You need a wood_pickaxe or a better tool 
            to mine stone."
            success = False
            suggestion = "Consider crafting or acquiring a wood_pickaxe or a better 
            tool to mine stone."
    else:
        feedback = "Action executed successfully."
        success = True
        suggestion = ""
    return feedback, success, suggestion
\end{lstlisting}
\end{adjustwidth}
\newpage
\begin{adjustwidth}{\dimexpr(\textwidth-0.95\textwidth)/2}{}
\lstset{language=python}
\begin{lstlisting}[linewidth=0.95\textwidth]
def Rule_5_mine(state, action, knowledge_graph):
    # Extract action details
    action_name = action.get("action_name")
    block_name = action.get("args", {}).get("block_name")

    # Initialize feedback, success, and suggestion
    feedback = ""
    success = True
    suggestion = ""

    # Check if the action is 'mine' and block_name is 'plant'
    if action_name == "mine" and block_name == "plant":
        feedback = "Action failed: You cannot mine a plant."
        success = False
        suggestion = "Consider mining other resources like 'tree' or 'stone' 
        instead of 'plant'."

    else:
        feedback = "Action executed successfully."

    return feedback, success, suggestion
\end{lstlisting}
\end{adjustwidth}

\newpage
\subsection{Action Rules in ALFWorld}

\begin{center}
    \textbf{Action Rules for ALFWorld}
\end{center}
\begin{adjustwidth}{\dimexpr(\textwidth-0.95\textwidth)/2}{}
\lstset{language=}
\begin{lstlisting}[linewidth=0.95\textwidth]
Rule 1: For action 'take', if the item is not present in the specified location, 
the action will fail; Checking Method: Check if the 'obj' in 'action' is not 
listed under 'items_in_locations' for the 'source' location in 'inital_state'. 
Rule 2: For action 'take', if the agent's hand is already holding an item, the 
action will fail; Checking Method: Check if 'item_in_hand' in 'inital_state' 
has a non-null 'item_name'. 
Rule 3: For action 'take', if the agent is not at the specified location, the 
action will fail; Checking Method: Check if 'current_position' in 
'inital_state' does not match the 'source' location in 'action'. 
Rule 2: For action 'open', if the target location is not in the list of 
reachable locations, the action will fail; Checking Method: Check if 
'action.args.target' is not in 'reachable_locations'. 
Rule 7: For action 'open', if the target location is a fridge and the 
current position is not the same as the target location, the action will 
fail; Checking Method: Check if 'action.args.target' is 'fridge' and 
'current_position.location_name' is not equal to 'action.args.target'. 
Rule 10: For action 'open', if the target location is a drawer and the 
current position is not the same as the target location, the action will 
fail; Checking Method: Check if 'action.args.target' starts with 'drawer'
and 'current_position.location_name' is not equal to 'action.args.target'. 
Rule 3: For action 'go to', if the item specified by 'item_name' in 
'item_in_hand' is None and the action's 'target' in the Scene Graph is 
not 'Unexplored' or does not contain the state's 'target_item' in the 
list, the action will fail; Checking Method: Check if 'item_in_hand.
item_name' is None, and validate if the 'target' in 'action' is either 
not labeled 'Unexplored' or does not include the 'target_item' in the 
Scene Graph. 
Rule 2: For action 'heat', if the object to be heated is not in the
tool, the action will fail; Checking Method: Check if the 'item_in_hand'
matches the object to be heated and is not placed in the tool location. 
Rule 2: For action 'put', if the item to be put is not in hand, the 
action will fail; Checking Method: Check if 'item_name' in 'item_in_hand' 
of the 'inital_state' matches the 'obj' in action args. 
Rule 4: For action 'put', if the current position is not the same as 
the target location, the action will fail; Checking Method: Check if 
'location_name' in 'current_position' of the 'inital_state' matches 
the 'target' in action args. 
Rule 2: For action 'use', if the object is not in the current location, 
the action will fail; Checking Method: Check if the object specified 
in the action is not present in the 'items_in_locations' of the 
'current_position' in the initial state. 
Rule 4: For action clean, if the object to be cleaned is not in hand, 
the action will fail; Checking Method: Check if the 'item_in_hand' 
matches the object specified in the action. 




    
\end{lstlisting}
\end{adjustwidth}

\newpage
\begin{center}
    \textbf{Code Rules for ALFWorld}
\end{center}
\begin{adjustwidth}{\dimexpr(\textwidth-0.95\textwidth)/2}{}
\lstset{language=}
\begin{lstlisting}[linewidth=0.95\textwidth]

def Rule_1_take(state, action, scene_graph):
    # Extract action details
    action_name = action.get("name")
    action_args = action.get("args", {})
    # Check if the action is 'take'
    if action_name == "take":
        obj = action_args.get("obj")
        source = action_args.get("source")
        # Check if the object is present in the specified source location
        items_in_source = state.get("items_in_locations", {}).get(source, [])
        if obj not in items_in_source:
            feedback = f"Action failed: {obj} is not present in {source}."
            success = False
            suggestion = f"Check the items in {source} and try taking an available item."
            return feedback, success, suggestion
    # If the action is not 'take', consider it successful
    feedback = "Action executed successfully."
    success = True
    suggestion = ""
    return feedback, success, suggestion
\end{lstlisting}
\end{adjustwidth}
\begin{adjustwidth}{\dimexpr(\textwidth-0.95\textwidth)/2}{}
\lstset{language=python}
\begin{lstlisting}[linewidth=0.95\textwidth]
def Rule_2_take(state, action, scene_graph):
    # Extract the action name
    action_name = action.get("name")
    # Check if the action is 'take'
    if action_name == "take":
        # Check if the agent's hand is already holding an item
        item_in_hand = state.get("item_in_hand", {}).get("item_name")
        if item_in_hand:
            feedback = "Action failed: The agent is already holding an item."
            success = False
            suggestion = "Put down the current item before taking another one."
            return feedback, success, suggestion
    # If the action is not 'take', consider it successful by default
    feedback = "Action executed successfully."
    success = True
    suggestion = ""
    return feedback, success, suggestion
\end{lstlisting}
\end{adjustwidth}
\begin{adjustwidth}{\dimexpr(\textwidth-0.95\textwidth)/2}{}
\lstset{language=python}
\begin{lstlisting}[linewidth=0.95\textwidth]
def Rule_3_take(state, action, scene_graph):
    if action['name'] == 'take':
        current_location = state['current_position']['location_name']
        source_location = action['args']['source']
        if current_location != source_location:
            feedback = "Action failed: You are not at the specified location to take the item."
            success = False
            suggestion = f"Move to {source_location} before attempting to take the item."
            return feedback, success, suggestion
    # If the action is not 'take', consider it successful by default
    feedback = "Action executed successfully."
    success = True
    suggestion = ""
    return feedback, success, suggestion
\end{lstlisting}
\end{adjustwidth}
\newpage
\begin{adjustwidth}{\dimexpr(\textwidth-0.95\textwidth)/2}{}
\lstset{language=python}
\begin{lstlisting}[linewidth=0.95\textwidth]
def Rule_2_open(state, action, scene_graph):
    if action['name'] == 'open':
        target = action['args']['target']
        if target not in state['reachable_locations']:
            feedback = f"Action failed: The target location '{target}' is not reachable."
            success = False
            suggestion = f"Try moving closer to '{target}' before attempting to open it."
            return feedback, success, suggestion
    # If the action is not 'open', consider it successful by default
    feedback = "Action executed successfully."
    success = True
    suggestion = ""
    return feedback, success, suggestion
\end{lstlisting}
\end{adjustwidth}
\begin{adjustwidth}{\dimexpr(\textwidth-0.95\textwidth)/2}{}
\lstset{language=python}
\begin{lstlisting}[linewidth=0.95\textwidth]
def Rule_7_open(state, action, scene_graph):
    # Extract necessary information from state and action
    current_position = state["current_position"]["location_name"]
    action_name = action["name"]
    target = action["args"].get("target", "")

    # Check if the action is 'open' and the target is a fridge
    if action_name == "open" and "fridge" in target:
        # Check if the current position is not the same as the target location
        if current_position != target:
            feedback = "Action failed: You must be at the fridge to open it."
            success = False
            suggestion = f"Move to {target} before trying to open it."
            return feedback, success, suggestion

    # If the action is not 'open' or the rule does not apply, consider it successful
    feedback = "Action executed successfully."
    success = True
    suggestion = ""
    return feedback, success, suggestion
\end{lstlisting}
\end{adjustwidth}
\begin{adjustwidth}{\dimexpr(\textwidth-0.95\textwidth)/2}{}
\lstset{language=python}
\begin{lstlisting}[linewidth=0.95\textwidth]
def Rule_10_open(state, action, scene_graph):
    # Extract necessary information from state and action
    current_position = state["current_position"]["location_name"]
    action_name = action["name"]
    target_location = action["args"].get("target", "")

    # Check if the action is 'open' and the target is a drawer
    if action_name == "open" and target_location.startswith("drawer"):
        # Check if the current position is not the same as the target location
        if current_position != target_location:
            feedback = "Action failed: You must be at the drawer to open it."
            success = False
            suggestion = f"Move to {target_location} before trying to open it."
            return feedback, success, suggestion

    # If the action is not 'open' or the rule does not apply, consider it successful
    feedback = "Action executed successfully."
    success = True
    suggestion = ""
    return feedback, success, suggestion
\end{lstlisting}
\end{adjustwidth}
\newpage
\begin{adjustwidth}{\dimexpr(\textwidth-0.95\textwidth)/2}{}
\lstset{language=python}
\begin{lstlisting}[linewidth=0.95\textwidth]
def Rule_3_go_to(state, action, scene_graph):
    # Extract necessary information from state and action
    item_in_hand = state["item_in_hand"]["item_name"]
    target_item = state["target_item"]
    action_name = action["name"]

    # Check if the action is 'go_to'
    if action_name == "go to":
        target_location = action["args"]["target"]
        # Check if item in hand is None
        if item_in_hand is None:
            # Check if the target location is not 'Unexplored' and does not contain the target item
            if target_location in scene_graph["locations"]:
                location_items = scene_graph["locations"][target_location]
                if "Unexplored" not in location_items and not any(target_item in item for item in location_items):
                    # Search for a location containing the target item
                    location_with_target_item = None
                    unexplored_locations = []

                    for location, items in scene_graph["locations"].items():
                        if any(target_item in item for item in items):
                            location_with_target_item = location
                            break  # Stop searching once a location with the target item is found
                        if "Unexplored" in items:
                            unexplored_locations.append(location)

                    # Prepare suggestion string
                    if location_with_target_item:
                        suggestion = f"According to scene graph. Please go to the location containing the target item: {location_with_target_item}."
                    elif unexplored_locations:
                        suggestion = (
                            "According to scene graph. "
                            "Please explore the following unexplored locations: "
                            + ", ".join(unexplored_locations) + "."
                        )
                    else:
                        suggestion = "No valid locations found that contain the target item or are unexplored."

                    feedback = f"Action failed: There is no {target_item} in {target_location} "
                    success = False
                    return feedback, success, suggestion

    # If the action is not 'go_to' or the conditions are not met, consider it successful
    feedback = "Action executed successfully."
    success = True
    suggestion = ""
    return feedback, success, suggestion
\end{lstlisting}
\end{adjustwidth}
\newpage
\begin{adjustwidth}{\dimexpr(\textwidth-0.95\textwidth)/2}{}
\lstset{language=python}
\begin{lstlisting}[linewidth=0.95\textwidth]
def Rule_2_heat(state, action, scene_graph):
    # Extract necessary information from the state and action
    action_name = action.get("name")
    action_args = action.get("args", {})
    obj_to_heat = action_args.get("obj")
    tool = action_args.get("tool")
    # Check if the action is 'heat'
    if action_name == "heat":
        # Check if the object to be heated is in hand and not in the tool location
        item_in_hand = state.get("item_in_hand", {}).get("item_name")
        current_position = state.get("current_position", {}).get("location_name")
        if item_in_hand == obj_to_heat and current_position != tool:
            feedback = f"Failed to heat {obj_to_heat}. It must be placed in {tool} to be heated."
            success = False
            suggestion = f"Place {obj_to_heat} in {tool} before heating."
            return feedback, success, suggestion
    # If the action is not 'heat', consider it successful
    feedback = "Action executed successfully."
    success = True
    suggestion = ""
    return feedback, success, suggestion
\end{lstlisting}
\end{adjustwidth}
\begin{adjustwidth}{\dimexpr(\textwidth-0.95\textwidth)/2}{}
\lstset{language=python}
\begin{lstlisting}[linewidth=0.95\textwidth]
def Rule_2_put(state, action, scene_graph):
    if action['name'] == 'put':
        obj_to_put = action['args']['obj']
        item_in_hand = state['item_in_hand']['item_name']
        if obj_to_put != item_in_hand:
            feedback = f"Action failed: {obj_to_put} is not in hand."
            success = False
            suggestion = f"Ensure you have {obj_to_put} in hand before attempting to put it."
            return feedback, success, suggestion
    # If the action is not 'put', consider it successful by default
    feedback = "Action executed successfully."
    success = True
    suggestion = ""
    return feedback, success, suggestion
\end{lstlisting}
\end{adjustwidth}
\begin{adjustwidth}{\dimexpr(\textwidth-0.95\textwidth)/2}{}
\lstset{language=python}
\begin{lstlisting}[linewidth=0.95\textwidth]
def Rule_4_put(state, action, scene_graph):
    if action['name'] == 'put':
        current_location = state['current_position']['location_name']
        target_location = action['args']['target']
        if current_location != target_location:
            feedback = "Action failed: You must be at the target location to put the item."
            success = False
            suggestion = f"Move to {target_location} before attempting to put the item."
            return feedback, success, suggestion
    # If the action is not 'put', consider it successful by default
    feedback = "Action executed successfully."
    success = True
    suggestion = ""
    return feedback, success, suggestion
\end{lstlisting}
\end{adjustwidth}
\newpage
\begin{adjustwidth}{\dimexpr(\textwidth-0.95\textwidth)/2}{}
\lstset{language=python}
\begin{lstlisting}[linewidth=0.95\textwidth]
def Rule_2_use(state, action, scene_graph):
    if action['name'] == 'use':
        obj = action['args']['obj']
        current_location = state['current_position']['location_name']
        # Check if the object is in the current location
        if obj not in state['items_in_locations'].get(current_location, []):
            feedback = f"Action failed: {obj} is not in the current location {current_location}."
            success = False
            suggestion = f"Move to the location where {obj} is present or bring {obj} to the current location."
            return feedback, success, suggestion
    # If the action is not 'use', consider it successful
    feedback = "Action executed successfully."
    success = True
    suggestion = ""
    return feedback, success, suggestion
\end{lstlisting}
\end{adjustwidth}
\begin{adjustwidth}{\dimexpr(\textwidth-0.95\textwidth)/2}{}
\lstset{language=python}
\begin{lstlisting}[linewidth=0.95\textwidth]
def Rule_4_clean(state, action, scene_graph):
    # Extract the action name and arguments
    action_name = action.get("name")
    action_args = action.get("args", {})
    
    # Check if the action is 'clean'
    if action_name == "clean":
        # Get the object to be cleaned from the action arguments
        obj_to_clean = action_args.get("obj")
        
        # Get the item currently in hand
        item_in_hand = state.get("item_in_hand", {}).get("item_name")
        
        # Check if the object to be cleaned is in hand
        if obj_to_clean != item_in_hand:
            feedback = f"Action failed: {obj_to_clean} is not in hand."
            success = False
            suggestion = f"Please ensure {obj_to_clean} is in hand before cleaning."
            return feedback, success, suggestion
    
    # If the action is not 'clean', consider it successful
    feedback = "Action executed successfully."
    success = True
    suggestion = ""
    return feedback, success, suggestion


\end{lstlisting}
\end{adjustwidth}

\section{Experiment Details}
\label{sec:Experiment Details}

\subsection{Mars}
\label{sec:Mars Experiment Details}

\textbf{Task Details.} 
Mars is an open-world environment designed for situated inductive reasoning, where agents must actively interact with their surroundings, induce generalizable rules, and apply them to achieve specific goals. Unlike traditional environments that rely on pre-existing commonsense knowledge, Mars introduces counter-commonsense mechanisms by modifying terrain distributions, survival settings, and task dependencies.

The agent’s goal is to unlock various achievements, such as:
\begin{itemize}
    \item Collecting Achievements: Agents gather essential resources, though collection methods may be counterintuitive (e.g., Collect Coal, Collect Diamond, Collect Drink, Collect Iron, Collect Sapling, Collect Stone, Collect Wood)
    \item Crafting Achievements: Agents create tools and weapons, but recipes may be altered (e.g., Make Wooden Pickaxe, Make Stone Pickaxe, Make Iron Pickaxe, Make Wooden Sword, Make Stone Sword, Make Iron Sword)
    \item Placing Achievements: Agents construct objects with modified material requirements (e.g., Place Table, Place Furnace, Place Plant, Place Stone)
    \item Survival and Combat Achievements: Tasks involve combat and survival, but behaviors may be altered (e.g., Kill Skeleton, Kill Zombie, Kill Cow, Eat Plant, Wake Up )
\end{itemize}

However, achieving these goals requires adaptive reasoning since the default assumptions about item acquisition and crafting may no longer hold. For example:
Mars introduces counter-commonsense modifications to challenge agents' reliance on prior knowledge. These modifications fall into three categories: Terrain, Survival Settings, and Task Dependencies, which can be combined to create diverse worlds.
\begin{itemize}
    \item Terrain Modifications: 
    \begin{itemize}
        \item Distribution: Resources appear in unexpected locations (e.g., diamonds in sand, coal near grass). 
        \item Effects: Terrain properties change (e.g., lava is safe, grass is harmful, mining stone gives wood).
    \end{itemize}
    \item Survival Setting Changes: 
    \begin{itemize}
        \item Entity Behavior: Cows may be aggressive, zombies passive, skeletons use melee instead of bows.
        \item Food and Health: Eating cows might reduce health, drinking lava could restore it.
    \end{itemize}
    \item Task Dependency Alterations: 
    \begin{itemize}
        \item Resource Collection: Mining may yield unexpected materials (e.g., trees drop iron, stone gives coal).
        \item Crafting Changes: Items require different materials (e.g., tables need diamonds, pickaxes need iron).
        \item Placement Rules: Placing objects may consume extra resources (e.g., tables require two diamonds).
    \end{itemize}
\end{itemize}
By combining these modifications, Mars forces agents to learn dynamically, making pre-stored knowledge unreliable and requiring real-time adaptation.

Each episode in Mars generates a unique $64 * 64$ grid-based world where agents operate under partial observability ($7 * 9$ grid view). The environment ensures task achievability by maintaining resource balance, enforcing supply constraints, and allowing procedural rule induction.

\textbf{Baselines.}
We compare our method against ReAct~\citep{yao2022react}, which interleaves reasoning and actions; Reflexion~\citep{shinn2023reflexion}, an extension of ReAct with self-reflection; Skill Library~\citep{tang2024mars}, an adaptation of JARVIS-1~\citep{wang2024jarvis} and Voyager~\citep{wang2023voyager} that stores successful plans for in-context learning; and IfR~\citep{tang2024mars}, which extends Skill Library with an induction-from-reflection module that derives and stores game rules for adaptive decision-making. These methods' planning framework and components are shown in Table~\ref{tab:planning framework}.

\begin{table}[h]
\caption{Comparison of baselines' planning framework with different components.}
\label{tab:planning framework}
\begin{center}
\begin{small}
\begin{sc}
\vspace{-1em}
\renewcommand\arraystretch{0.85}
\resizebox{\linewidth}{!}{
    \begin{tabular}{lcccccc}
    \toprule
    \textbf{Method}                     & reasoning                 & Experience Reflector  & memory/skill library             &        rules    & knowledge graph & scene graph \\ \midrule
    ReAct~\citep{yao2022react}          &  \checkmark               &                       &                                  &                 &                 &               \\
    Reflexion~\citep{shinn2023reflexion}&  \checkmark               &  \checkmark           &                                  &                 &                 &                \\
    Skill Library~\citep{wang2024jarvis}& \checkmark                &  \checkmark           &    \checkmark                    &                 &                 &              \\
    IfR~\citep{tang2024mars}            & \checkmark                &  \checkmark           &    \checkmark                    &   \checkmark    &                 &               \\
    \textbf{\ours (ours)}              &  \checkmark               &  \checkmark           &                                  &   \checkmark    &   \checkmark    &   \checkmark    \\ \bottomrule
    \end{tabular}
}
\end{sc}
\end{small}
\end{center}
\end{table}

\textbf{Method Setup.} 
We utilize GPT-4 as the backend for our method. Adopting the same experimental setup as Mars, we set the number of learning episodes to 5 and conduct 9 independent trials for evaluation. We employ a one-step MPC approach, where the world model assesses whether the agent's current action aligns with the environment's mechanisms based on its state information. The world model then provides feedback and suggestions to the agent, enabling it to refine its plan based on the state preceding the failed action and the received feedback.

\subsection{ALFWorld}
\label{sec:ALFWorld Experiment Details}

\textbf{Task Details.} ALFWorld is a virtual environment designed as a text-based simulation where agents perform tasks by interacting with a simulated household. The environment includes six distinct task types, each requiring the agent to accomplish a high-level objective, such as placing a cooled lettuce on a countertop. Agents use text commands to navigate and manipulate objects in the virtual space, for example, issuing instructions like "go to countertop 1," "take lettuce 1 from countertop 1," or "cool lettuce 1 with fridge 1." The visual observations from the agent’s point of view are converted into natural language descriptions before being delivered to the agent. The agent's state is represented by the cumulative history of these observations. Success is measured by the completion the specified task goal.

\textbf{Method Setup.} 
We conducted rule learning on the training set, with the resulting action rules presented in Appendix \ref{sec:Learned Rules}. 
We adopted a one-step MPC. This method evaluates whether the agent's current action aligns with the environment's dynamic patterns based on its state information.
Additionally, to enhance rule discovery, we developed scripts to convert the natural language dialogue history and action information into a structured JSON format, as illustrated in Appendix \ref{sec: ALFworld Environments' State Space and Action Space}. 
We utilize GPT-3.5-Instruct as our backbone model.

\subsection{Experiment Design for Effectiveness of NeuroSymbolic Learning}
\label{sec:Experiment Design for Effectiveness of Rule Learning}

We set the number of learning episodes to 5. After each episode, the model, equipped with latest learned symbolic knowledge or skill library, is tested on the testing set.

The cover rate quantifies the extent to which the rules derived from the neurosymbolic learning process address the LLM's failed predictions. 
Specifically, it represents the probability that mispredicted transitions by the LLM are correctly handled by the learned rules.

To assess the alignment between the LLM-based world model and the actual environment, 
we first identify transitions where the LLM fails to make accurate predictions. This is achieved by utilizing an unaligned LLM world model—one without any symbolic knowledge—to generate predictions for trajectories obtained from the test set. The discrepancies between the predicted observation \( \hat{o}_{t+1} \) and the actual observation \( o_{t+1} \) are compiled into a dataset of mispredicted transitions.
These mispredictions highlight areas where the LLM world model does not align with the environment's dynamics.

Subsequently, the learned rules at each iteration are evaluated against the mispredicted transitions dataset to determine their effectiveness in correcting these mispredictions.
If a rule successfully predicts the outcome of a previously mispredicted transition, it demonstrates that the rule effectively addresses the LLM's failure in that instance. The cover rate is then calculated as the ratio of correctly addressed mispredictions to the total number of mispredicted transitions:
\begin{equation}
    \text{Cover Rate} = \frac{\text{Number of Mispredictions Addressed by Rules}}{\text{Total Number of Mispredicted Transitions}}
\end{equation}
A higher cover rate indicates that the neurosymbolic learning process effectively enhances the alignment of the LLM world model with the environment, thereby improving the overall accuracy and reliability of the agent's planning.

\textbf{Determining Whether a Rule is Correct or Incorrect}
When a rule is active, if it makes an incorrect judgment—predicting success when the transition actually fails or vice versa—the rule is considered invalid and is removed from the rule set. Transitions where the rule is not applicable—referred to as "inactive" or "dormant"—are excluded from the evaluation process.

\section{Limitation and Future Work}
Currently, our neurosymbolic learning framework generates simple rules that primarily assess whether actions align with environment dynamics (i.e., rules for transitions). Future research should explore advanced reasoning methods that enable LLMs to derive more abstract rules, such as those governing entire planning processes.
Furthermore, many embodied environments exhibit stochastic dynamics, where actions have probabilistic outcomes. For example, resource gathering at night in Mars often fails due to hostile creatures but can sometimes succeed. Our current rule learning process cannot handle such randomness, typically classifying these scenarios as failures. Addressing this limitation by enabling rules to account for stochastic dynamics is a promising research direction, potentially leading to more accurate and reliable world models.

\end{document}